\documentclass{article}
\usepackage{iclr2027_conference,times}
\iclrfinalcopy 

\usepackage{amsmath,amssymb,bm}
\usepackage{booktabs}
\usepackage{array}
\usepackage{graphicx}
\usepackage{tikz}
\usetikzlibrary{arrows.meta,positioning,calc,fit,backgrounds,shapes.geometric}
\usepackage{xcolor}
\usepackage{hyperref}
\usepackage{url}
\usepackage{caption}
\newcommand{\primD}{\textsc{Prim-1D}}      
\newcommand{\mixD}{\textsc{Mix-1D}}        
\newcommand{\evalN}{\textsc{Eval-9}}       
\newcommand{\nsD}{\textsc{NS-2D}}          
\newcommand{\slib}[1]{$\mathcal{L}^{\mathrm{SL}}_{#1}$}   
\newcommand{\dlib}{$\mathcal{L}^{\mathrm{SSL}}$}          
\title{SPLICE: Sparse Physics Library Identification and Compositional Expansion}

\author{%
  Lucas Thil \\
  LIX, École Polytechnique \\
  IRT SystemX \\
  Palaiseau, France \\
  \texttt{thil@lix.polytechnique.fr} \\
  \And
  Jesse Read \\
  LIX, École Polytechnique \\
  Palaiseau, France \\
  \And
  Rim Kaddah \\
  IRT SystemX \\
  Palaiseau, France \\
  \And
  Guillaume Doquet \\
  Safran Tech \\
  Chateaufort, France \\
}

\begin{document}
\maketitle
\lhead{Preprint}        

\begin{abstract}

Neural operator splitting composes a library of neural operators, one per physical mechanism, into zero-shot simulators of new equations. Existing approaches, however, require these mechanisms and their labels to be specified in advance. We ask whether a pretrained library can instead be adapted and grown from unlabeled trajectories in which multiple mechanisms act simultaneously.
We introduce SPLICE, a framework that discovers operators without mechanism labels on mixture data.
The contributions of the candidate operators to each trajectory are inferred by solving a convex optimization problem. 
A calibrated test then assigns each discovered operator the mechanism it implements.
Failure cases are also detectable, when the library lacks the required physics or when mechanisms are insufficiently separable.
Across one and two dimensional systems, the discovered operators recover mechanisms from unlabeled mixtures and compose into zero-shot simulators of unseen equations; a frozen library can also grow an operator for physics absent from pretraining, without labels and without forgetting. On Navier-Stokes, the discovered operators compose about as accurately as their supervised counterparts.


\end{abstract}

\section{Introduction}
\label{sec:intro}

Numerical simulation of physical systems underpins science and engineering, yet each new system typically requires its own solver or its own trained model \citep{Raissi_Perdikaris_Karniadakis_2019,kochkov2021,2020arXiv201003409P}. Physical systems, however, share structure: many are driven by a small set of basic effects, such as transport and spreading, acting together. Neural operator splitting (NOS)~\citep{serrano2026test} exploits this by learning one neural operator per effect and combining them at test time, so that a single reusable library simulates equations never seen in training, instead of one model per system \citep{mccabe2023multiple,herde2024poseidon}.

Such libraries must be specified in advance. Each operator is trained on data in which its effect acts alone and is labeled, so adding an effect requires new labeled data. Real trajectories rarely come this way: instead, several unlabeled effects act at once. We therefore ask: \emph{can an operator library learn new operators from unlabeled trajectories in which several effects act simultaneously, and can we tell what each learned operator does?}

\begin{figure}[t]
\centering
\includegraphics[width=0.95\linewidth]{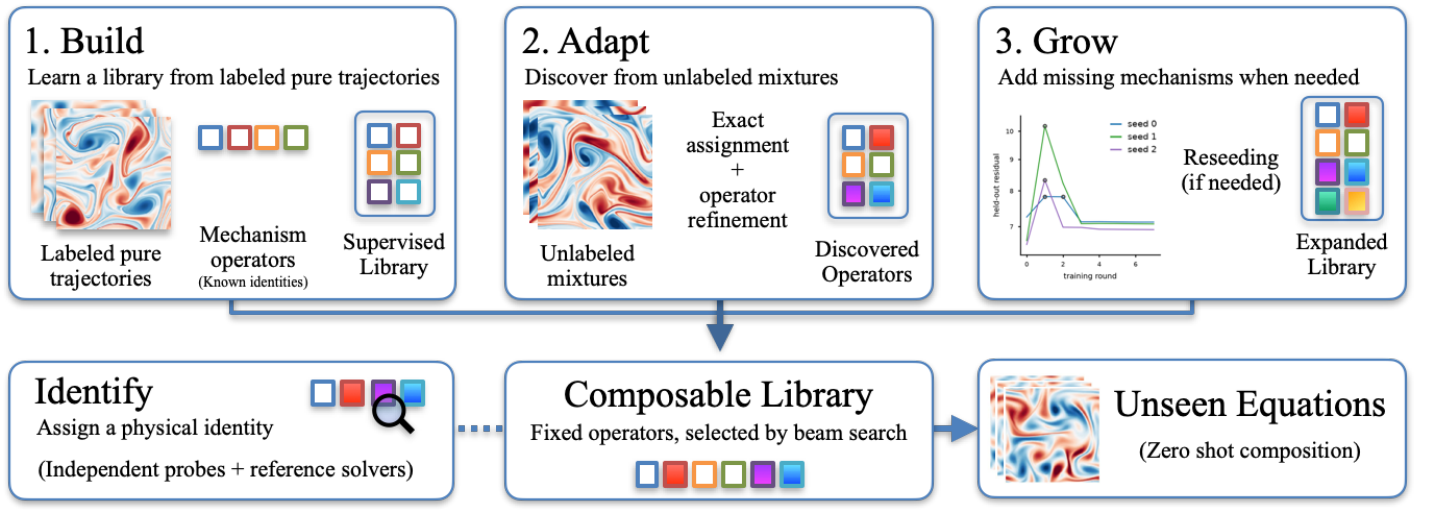}
\caption{\textbf{The SPLICE pipeline.}
\textit{Build} trains one operator per labeled effect and composes them, following \cite{serrano2026test}; \textit{Adapt}, \textit{Grow}, and \textit{Identify} are ours. \textit{Adapt} learns operators from unlabeled trajectories, and \textit{Grow} adds an operator when the data reveal dynamics the library cannot express. \textit{Identify} tests each learned operator against reference solvers used only for evaluation, never for training. At test time, a beam search selects which operators of the library to compose for each unseen equation.}
\label{fig:pipeline}
\end{figure}

We introduce SPLICE to answer both questions (Figure~\ref{fig:pipeline}). It starts from a library pretrained on a few labeled effects, adapts its operators on unlabeled data, and grows the library when the data reveal dynamics it cannot yet express; neither step uses effect labels. A final identification step then tests which physical effect each learned operator implements, by comparing its action with reference solvers that are never used for training.


In one and two dimensional systems, SPLICE recovers the hidden effects of unlabeled trajectories, including fluid flows in which no effect is ever observed alone, and the learned operators combine to simulate equations unseen during training. On Navier-Stokes, the discovered libraries compose about as accurately as their supervised counterparts, and can re-grow a missing library entry. The experiments also mark where the approach stops: effects with near-identical actions end up sharing one operator, an operator can be correctly identified yet fail to improve a particular combination (\S\ref{sec:res-negative}), 
and growth re-grows effects that the pretrained library can represent. 
An effect absent from pretraining can be added as a new operator, learned without labels from the windows the library reconstructs poorly, without forgetting and at a fifth of the compute of retraining, though retraining with replay remains more accurate over long rollouts.

\paragraph{Contributions.}
\begin{itemize}\setlength\itemsep{2pt}
  \item \textbf{A label-free method to adapt and grow operator libraries.} We introduce SPLICE, which adapts a pretrained operator library on unlabeled trajectories in which several effects act at once, and adds an operator when the data reveal dynamics the library cannot express.
  \item \textbf{A protocol to identify what each operator learned.} We propose an evaluation protocol, kept separate from training, that assigns each learned operator a physical effect by comparison with reference solvers, or reports that none can be assigned, so that a failing simulation points to the operator responsible.
  \item \textbf{Evidence that label-free libraries match supervised ones.} On 1D and 2D systems, including Navier--Stokes flows in which no effect ever acts alone, the adapted libraries recover the hidden effects and compose Navier--Stokes as accurately as supervised libraries.
\end{itemize}

\section{Related work}
\label{sec:related}

Neural operators such as the Fourier neural operator (FNO) \citep{li2020fourier} learn maps between function spaces \citep{lu2021learning,JMLR:v24:21-1524,koren2025merging}. Breadth is increasingly pursued through scale and conditioning: multi-physics pretraining \citep{mccabe2023multiple,herde2024poseidon,hao2024dpot,alkin2024universal}, in-context and adaptive conditioning \citep{serrano2024zebra,kassai2024boosting,Yang_2023}, PDE-conditional transformers \citep{pmlr-v267-zhou25r}, and continually grown component libraries \citep{2023arXiv231018885T,mendez2021lifelong}, with labeled per-physics data in the physical case. All reach coverage through scale, conditioning, or adaptation rather than through a library of independently identifiable mechanism operators. 

Classical operator splitting \citep{strang1968} offers a different route: compose solvers for the individual terms of the governing equation. NOS \citep{serrano2026test} makes these solvers learned and searches over their compositions at test time; we build on it and ask how such a library can be discovered and grown when mechanism identities are unavailable. Concurrent work, HyCOP \citep{zhao2026hycophybridcompositionoperators}, also builds a dictionary of composable PDE primitives, but from supervised residuals with a learned composition schedule, where we discover operators without mechanism labels under the fixed beam-searched Strang composition of NOS.

SINDy \citep{brunton2021modern}, PDE-FIND \citep{rudy2017data}, and their weak-form and ensemble variants \citep{fasel2022ensemble} recover governing equations by sparse regression over candidate terms, and neural variants learn the terms themselves \citep{long2018pde,both2021deepmod,sindyae2019} or distill them from trained models \citep{Cranmer_SanchezGonzalez_Battaglia_Xu_Cranmer_Spergel_Ho_2020}. DMD and Koopman methods fit global evolution operators \citep{schmid2010dynamic,brunton2021modern}, operator inference assumes the governing structure \citep{peherstorfer2016data}, and switching state-space models find regimes without naming them \citep{pmlr-v54-linderman17a}. These methods recover equations or components within a specified representation; few yield reusable per-mechanism solvers whose physical identity is established by an external test. PDE-FIND also serves as a baseline in \S\ref{sec:res-ssl}.

LEADS \citep{yin2021leads}, CoDA \citep{kirchmeyer2022generalizing}, and DyAd \citep{wang2022meta} decompose unlabeled dynamics into shared or environment-specific components, but the components are not assigned a physical identity, and independent-mechanism learning discovers competing modules without naming them \citep{pmlr-v80-parascandolo18a}. Unsupervised identifiability is known to require external structure \citep{pmlr-v97-locatello19a}; our setting supplies it after the fact:  a discovered operator is graded against independently specified reference dynamics, which makes discovery a falsifiable identification problem rather than only a decomposition.

Closest to our evaluation design, \citet{garrido2023self} grade self-supervised representations against transformation parameters known only to the evaluator, as discovered operators are graded here against a mechanism basis withheld from training. The Well \citep{ohana2024well} supplies our external transfer data (\S\ref{sec:res-transfer}); its coupled multi-mechanism systems suit this setting more directly than largely single-equation suites \citep{Takamoto_Praditia_Leiteritz_MacKinlay_Alesiani_Pflüger_Niepert_2022}.
\section{Approach}
\label{sec:approach}

\subsection{Background: neural solver operators and their composition}
\label{sec:background}

A trajectory is a sequence of fields $u_{0:T}$, with $u_t \in \mathbb{R}^{C\times N}$ in 1D and $u_t \in \mathbb{R}^{C\times N\times N}$ in 2D, where $C$ is the number of physical channels (e.g.\ velocity components) and $N$ the number of grid points per dimension; Table~\ref{tab:notation} (appendix) collects all symbols. We assume the dynamics are a sum of a few effects, $\partial_t u = \sum_m c_m A_m(u)$, where each $A_m$ is a primitive mechanism (one of the basic effects of \S\ref{sec:intro}), such as advection (transport) or diffusion (spreading), and $c_m \ge 0$ is its strength.

\begin{figure}[t]
\centering
\resizebox{0.9\linewidth}{!}{%
\begin{tikzpicture}[
  font=\small,>=Latex,
  node distance=4mm and 6mm,
  data/.style={draw,rounded corners=2pt,fill=blue!8,align=center,inner sep=2.5pt,minimum height=8mm,minimum width=14mm},
  mod/.style={draw,fill=black!7,align=center,inner sep=2.5pt,minimum height=8mm,minimum width=14mm},
  lib/.style={draw,fill=green!10,align=center,inner sep=3pt,minimum height=10mm,minimum width=26mm},
  note/.style={font=\footnotesize\itshape,text=black!55},
  hd/.style={font=\bfseries},
  ar/.style={->,thick},
  aro/.style={->,thick,black!55},
]
\node[data] (ctx) {context\\$u_{t-L:t}$};
\node[mod,right=of ctx] (enc) {encoder\\$E$};
\node[data,right=of enc] (z) {code $\theta$};
\node[mod,right=of z] (gen) {hyper-net\\$G$};
\node[mod,right=of gen] (f) {vector field\\$f_\theta$};
\node[mod,right=of f] (rk) {RK4\\step};
\node[data,right=of rk] (u1) {$\hat u_{t+1}$};
\foreach \a/\b in {ctx/enc,enc/z,z/gen,gen/f,f/rk,rk/u1} \draw[ar](\a)--(\b);
\node[hd,above=1mm of ctx.north west,anchor=south west] {(a) One operator $\Phi_\theta$: a region of code space drives a learned time-stepper};
\node[lib,below=11mm of z] (lib) {library\\$\mathcal{L}=\{\Phi_1,\dots,\Phi_M\}$};
\draw[aro] (z.south) -- (lib.north) node[pos=0.22,right,note]{one code per mechanism};
\node[data,left=10mm of lib] (pure) {labeled pure-\\mechanism data};
\draw[ar](pure)--node[above,note]{train}(lib);
\node[hd,above=1mm of pure.north] {(b) Build};
\node[mod,right=10mm of lib] (comp) {beam Strang\\composition};
\node[data,right=of comp] (eq) {unseen\\equation};
\draw[ar](lib)--(comp); \draw[ar](comp)--(eq);
\node[hd,above=1mm of comp.north] {(d) Compose};
\node[data,below=9mm of pure] (mix) {unlabeled\\mixtures};
\node[mod,right=of mix] (w) {w-step\\convex NNLS};
\node[mod,right=8mm of w] (phi) {$\phi$-step\\refine};
\draw[ar](mix)--(w);
\draw[ar,transform canvas={yshift=1.5mm}] (w)--(phi);
\draw[ar,transform canvas={yshift=-1.5mm}] (phi)--(w);
\node[mod,below=4mm of w] (rs) {reseed worst-residual window};
\draw[ar](phi.south)|-(rs.east);
\draw[ar](rs.west)-|(mix.south);
\draw[aro](phi.north) to[out=90,in=-50] node[pos=0.45,right,note]{add discovered} (lib.south east);
\node[hd,above=1mm of mix.north west,anchor=south west] {(c) Adapt \& Grow (label-free)};
\node[mod,right=20mm of phi] (fp) {$d_m(\hat\Phi)$ vs.\ reference\\solvers and null};
\node[data,right=of fp] (idn) {mechanism\\$m$ or none};
\node[data,below=4mm of fp] (probe) {probe set $\mathcal{B}$};
\draw[ar](probe)--(fp); \draw[ar](fp)--(idn);
\draw[aro](phi.east)--node[above,note]{discovered $\hat\Phi$}(fp.west);
\node[hd,above=1mm of fp.north west,anchor=south west] {(e) Identify};
\end{tikzpicture}}
\caption{The shared architecture and where each stage acts. \textbf{(a)} Every library entry is one learned time-stepper, in which an encoder maps a context window to a code $\theta$, a hypernetwork turns $\theta$ into the weights of a vector field $f_\theta$, and RK4 advances one step, so one region of code space is one operator. \textbf{(b)} \emph{Build} trains the library on labeled pure-mechanism trajectories. \textbf{(c)} \emph{Adapt and Grow} alternate a convex assignment (w-step) with operator refinement ($\phi$-step) on unlabeled mixtures, and reseed an operator from the worst-reconstructed window. \textbf{(d)} \emph{Compose} beam-searches Strang compositions of library operators. \textbf{(e)} \emph{Identify} grades each discovered operator against reference solvers on a fixed probe set. Blue nodes are data; grey nodes are learned modules.}
\label{fig:arch}
\end{figure}
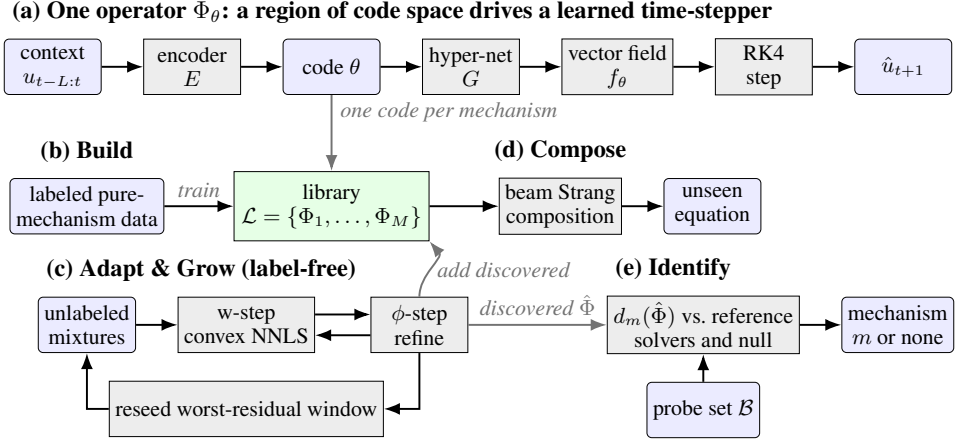

Following \citet{serrano2026test}, a single neural network represents every operator. An encoder $E$ reads a context window of $L$ frames and outputs a latent code $\theta = E(u_{t-L:t}) \in \mathbb{R}^{d}$. A hypernetwork maps $\theta$ to the weights of a vector field $f_\theta$, and one step integrates this field with the classical fourth-order Runge--Kutta scheme (RK4), a neural ODE step \citep{Chen_Rubanova_Bettencourt_Duvenaud_2018}:
\begin{equation}
\Phi_\theta(u_t) = \mathrm{RK4}[f_\theta](u_t) \approx u_{t+1}.
\label{eq:step}
\end{equation}
Trained on labeled single-mechanism trajectories, the code space organizes by mechanism, so each region of $\theta$ space acts as one operator. The library $\mathcal{L} = \{\Phi_{\theta_1}, \ldots, \Phi_{\theta_M}\}$ holds one operator per mechanism, $M$ being the number of mechanisms in the labeled data (Figure~\ref{fig:arch}).

To simulate an equation with several active mechanisms, operators are applied in sequence within each time step (Strang splitting: $\Phi_A^{1/2} \circ \Phi_B \circ \Phi_A^{1/2}$, local error $O(\Delta t^3)$). Which operators to combine is chosen at test time by beam search, a pruned search that keeps the few best partial compositions as it extends them and scores each by rollout error. We use this machinery as in \citet{serrano2026test}: an unseen equation can be simulated when its mechanisms are represented in the library.

\subsection{Build: learning a library of mechanism operators}
\label{sec:method-corpus}

\primD{} extends the two and three family settings of NOS to eight primitive families with dense coefficient coverage. We study corpus design and solver fidelity primarily in 1D, where exact spectral solvers make labeled data inexpensive to generate and multi seed studies practical. Before an operator is used in library, we require stable rollout on trajectories generated by its own mechanism.

\subsection{Adapt: discovering operators from unlabeled mixtures}
\label{sec:method-ssl}

We now observe trajectories whose dynamics are a sum of mechanisms as in Section~\ref{sec:background} with one or two active mechanisms 
and random coefficients, but with no label saying which mechanisms are active. The goal is to learn a set of \emph{discovered} operators $\{\hat\Phi_1,\ldots,\hat\Phi_K\}$, 
each of the form in Eq.~\eqref{eq:step} with its own learnable code fed to the frozen pretrained hypernetwork \citep[as task embeddings in][]{von2019continual}, together with
per trajectory assignments indicating which discovered operators contribute to each trajectory. Throughout, hatted symbols denote discovered quantities, $k$ indexes discovered operators, and $m$ is reserved for physical mechanisms; $K$ need not equal the number of supervised families $M$ and is chosen without labels (predictive $K$, below). Both discovery losses are defined only from observed future frames. The physical mechanism basis is withheld from discovery and used only for post hoc identification in \S\ref{sec:protocol}. We denote the resulting discovered library by \dlib{}.

\paragraph{Exact assignment by convex optimization.}
Because the active mechanisms are unobserved, discovery alternates between inferring which discovered operators contribute to each trajectory and refining them given those assignments. For fixed discovered operators, the inference step requires no learned component: because the dynamics are additive, the nonnegative mixture weights are obtained by a 
least squares problem solved to optimality for each trajectory. The joint problem over discovered operators and assignments remains nonconvex. Learned alternatives to this assignment step recover less and are compared in Appendix~\ref{app:amortized}. Let $z_t$ denote the frame $u_t$ standardized by its window mean and scale, and $\Delta z_t = z_{t+1} - z_t$ the resulting one step increment, a finite difference estimate of $\partial_t u$ at time $t$. The alternating procedure is
\begin{align}
\textbf{$w$-step (exact, convex):}\quad
w^{\star}
&= \arg\min_{w \ge 0}
\left\|
\sum_{k=1}^{K} w_k \hat f_k(z_t) - \Delta z_t
\right\|_2^2
+ \lambda \big(\textstyle\sum_k w_k\big)^2,
\label{eq:wstep}\\
\textbf{$\phi$-step (gradient):}\quad
\min_{\hat\phi}\quad
&\sum_{h \in \{1,4\}}
\left\|
\hat\Phi^{(h)}_{\hat\phi,w^{\star}}(u_t) - u_{t+h}
\right\|_2^2.
\label{eq:phistep}
\end{align}
Here $\|\cdot\|_2$ is the Euclidean norm over grid points, $w \in \mathbb{R}^K_{\ge 0}$ are per trajectory mixture weights, $\hat\phi$ collects the codes of the discovered operators, and
$\hat\Phi^{(h)}_{\hat\phi,w}$ rolls out the weight mixed field for $h$ steps from frame $u_t$. The w-step is analogous to sparse regression in SINDy \citep{brunton2016discovering}, with learned operators replacing symbolic terms. The multi horizon $\phi$-step refines the discovered operators with the assignment fixed and exposes slower mechanisms that may contribute little to one step prediction. We use $\lambda = 10^{-3}$ throughout; the w-step stays an exact nonnegative least-squares problem, whose nonnegativity keeps $w$ sparse. The weights $w$ are assignment weights rather than estimates of the physical coefficients $c_m$: the discovered operators' actions enter the fit unnormalized, and $w_m$ tracks the true coefficient magnitude only for advection (Appendix~\ref{app:opdist}).

\paragraph{Label free model selection.}
Merging duplicate discovered operators pooled across runs and choosing among runs are based on held out prediction on unlabeled data \citep{arlot2010survey}. Structural clustering heuristics fail in this setting (Appendix~\ref{app:opdist}), while the physical identification metric would use withheld mechanism information and is therefore not used for model selection.

\subsection{Grow: expanding the library when residuals suggest missing structure}
\label{sec:method-grow}

Adaptation cannot introduce a mechanism absent from the initial library: it moves operator codes within the fixed pretrained network, and only for operators that some trajectory already selects.
We thus monitor the held out reconstruction residual of Eq.~\eqref{eq:wstep}: a persistently high residual signals dynamics the current library cannot express. When an operator's assignment weight stays near zero, it is \emph{reseeded}: among the $R$ unlabeled windows ranked worst by held out residual, each candidate window is encoded to initialize a new operator, the operator is refined by the same self supervised objective (Eqs.~\eqref{eq:wstep}--\eqref{eq:phistep}), and the restart with the lowest held out residual is kept (best of $R$). Growth therefore uses only quantities available without labels, and mechanism identities remain unknown until identification. The procedure is dead operator replacement from dictionary learning \citep{aharon2006k,mairal2010online} transposed to dynamical operators; the reseeding details and the leave-one-mechanism-out study are in Appendix~\ref{app:reseed}.

\paragraph{Growing a new operator.} Codes cannot reach a mechanism that the pretrained network cannot represent. In that case we freeze the library and add a small new operator network, trained on the windows that the frozen library reconstructs poorly (held out residual above the $95$th percentile of known data). Its weight is fixed to one during training and solved by Eq.~\eqref{eq:wstep} only at evaluation. Since no pretrained weight changes, nothing is forgotten. The new operator starts from a small random initialization, because from zero the weight step removes it before it can learn.

\subsection{Identifying discovered operators}
\label{sec:protocol}\label{sec:identify}

A discovered operator is evaluated against an external mechanism 
only after discovery. The question is if its action on independent probe fields agrees more closely with one reference mechanism than with competent alternatives. We use a fixed probe set $\mathcal{B}$ of synthetic fields, generated once independently of training, and compare each discovered operator's displacement with the one produced by each reference solver, where $\mathcal{S}_m$ denotes one step of the exact reference solver for mechanism $m$:
\begin{equation}
d_m(\hat\Phi)
=
\operatorname*{mean}_{p \in \mathcal{B}}
\frac{
\left\|
(\hat\Phi(p)-p) - (\mathcal{S}_m(p)-p)
\right\|_2
}{
\left\|
\mathcal{S}_m(p)-p
\right\|_2
}.
\label{eq:fingerprint}
\end{equation}
Here $d_m(\hat\Phi)$ measures disagreement between the discovered operator and mechanism $m$ on the probe set, normalized by the magnitude of the reference displacement. Small $d_m$ therefore indicates agreement with that mechanism's action.

We evaluate discovered libraries along three dimensions. \emph{(i) Mechanism identification:} the closest reference mechanism must be preferred by a predefined margin, and the discovered operator's distance to that mechanism must fall below a null threshold set from the distances of operators that do not implement it (Appendix~\ref{app:protocol}). \emph{(ii) Selection:} on held out unlabeled mixtures, the discovered library must recover active mechanisms above permutation chance. \emph{(iii) Utility:} the error of rollouts composed from the discovered operators on held-out trajectories, measured in the same simulation harness as the supervised library (lower is better). Mechanism identification and simulation utility are therefore evaluated separately: agreement with a reference mechanism does not by itself imply that the discovered operator is sufficiently accurate for long rollouts.

The probe set, thresholds, and decision criteria are fixed before the corresponding evaluation. We report the resulting identification outcomes together with the selection and utility measurements. This separation keeps the mechanism basis out of discovery while making the physical interpretation of a discovered operator testable against independently specified reference dynamics.

\section{Experimental setup}
\label{sec:setup}

\label{sec:data}

Table~\ref{tab:data} defines every dataset and library name. Supervised libraries ($\mathcal{L}^{\mathrm{SL}}$) are trained on labeled pure mechanism trajectories. Discovered libraries ($\mathcal{L}^{\mathrm{SSL}}$) are learned from unlabeled mixtures and warm start from a supervised library trained on labeled synthetic primitives, where labels are free by construction (\S\ref{sec:method-ssl}). We denote by \emph{label-free} the fact that no mechanism labels enter training or model selection, rather the mechanism basis is used only by the post-hoc identification protocol (\S\ref{sec:identify}). \evalN{} is a test-only suite of nine held-out named equations, each a two-to-four-term composition of \primD{} families with jittered coefficients (the equations and exact forms are in Appendix~\ref{app:data}). 2D experiments and ablations run at $128^2$, and the main Navier--Stokes results are repeated at $256^2$, the resolution NOS reports.

We use two architectures: the NOS architecture \citep{serrano2026test} and a second, independently implemented \emph{control architecture}, which tests implementation independence (\S\ref{sec:res-breadth}). A \emph{supervised operator} is one library entry: the operator of Eq.~\eqref{eq:step} with its code held in the code-space region associated with one mechanism. All operators in a library come from a single jointly trained model, one encoder and generator whose code space is carved into per-mechanism regions; no separate network is trained per mechanism (Figure~\ref{fig:arch}). Library instances are per-dimension (1D on \primD{}, 2D on \nsD{}) and never share weights. We use two versions of the NOS architecture: the \emph{NOS operator}, our reproduction without the vector-quantized codebook and in-context conditioning, which keeps each operator's identity fixed, and the \emph{full NOS method}, with both (Appendix~\ref{app:nos}).

Rollout quality is per-step-mean relative $L_2$ error (NRMSE) over an autoregressive rollout: 16 steps for \evalN{} and \textsc{Abl-2D}, and 10 steps for the \nsD{} test set, the horizon of NOS's released test trajectories.

\section{Results}
\label{sec:results}
We first establish the breadth and solver fidelity of the supervised library that discovery builds on (\S\ref{sec:res-breadth}, \S\ref{sec:res-quality}). We then test what unlabeled discovery recovers (\S\ref{sec:res-ssl}, \S\ref{sec:res-nsssl}), and where its main limitations are (\S\ref{sec:res-negative}); transfer to external benchmark data is in Appendix~\ref{sec:res-transfer}.


\subsection{Composition over a broad corpus allows generalization to nine unseen equations}
\label{sec:res-breadth}

Before evaluating discovery, we first establish that the supervised library is broad enough to support composition on equations not seen during training. Table~\ref{tab:battery} evaluates zero-shot composition on \evalN{} through one shared beam harness. Growing the corpus from two to eight primitive families divides the error of the control architecture by about two across the nine unseen equations. The NOS recipe libraries reach the same range ($0.096 \pm 0.005$ for the NOS operator and $0.107 \pm 0.009$ for the full NOS method) with no noticeable differences beyond run variance.
Here, corpus breadth is the main contributor to result quality, while the choice of operator architecture has a smaller effect.

A monolithic FNO \cite{li2020fourier} trained on the same corpus reaches $0.117 \pm 0.001$, while a parameter-matched monolith ($16.4$M parameters) reaches $0.113 \pm 0.002$. Thus, the improvement is not explained by parameter count alone. Rather, composition allows the library to reuse mechanism specific operators across equations. Figure~\ref{fig:1dcomp} (appendix) shows one composed example, mechanism by mechanism.


\begin{table}[t]
\centering\small
\caption{Zero-shot composition on \evalN{} (mean NRMSE over nine equations; one shared beam harness; all models supervised). \slib{8} rows are mean$\pm$sd over 3 seeds; the NOS rows are measured at 3 seeds (Appendix~\ref{app:repro}) and are indistinguishable from the control architecture within run-to-run variance. The monolithic rows roll out one network trained on the same corpus. \slib{2} and \slib{8} are supervised libraries trained on two (advection, diffusion) or eight mechanism families. The main observation is that generalizing to new unseen equations is made possible thanks to the library breadth (per-equation results in Table~\ref{tab:pereq}; architectures defined in \S\ref{sec:data}).}
\label{tab:battery}
\begin{tabular}{llc}
\toprule
Architecture & Library & Mean \\
\midrule
control architecture & \slib{2} & 0.190 \\
control architecture & \slib{8} & 0.096 $\pm$ 0.001 \\
NOS operator & \slib{8} & 0.096 $\pm$ 0.005 \\
full NOS method & \slib{8} & 0.107 $\pm$ 0.009 \\
\midrule
monolithic FNO, no composition (2.2M) & trained on \primD{} & 0.117 $\pm$ 0.001 \\
monolithic FNO, parameter-matched (16.4M) & trained on \primD{} & 0.113 $\pm$ 0.002 \\
\bottomrule
\end{tabular}
\end{table}

\subsection{Solver fidelity dominates long horizon accuracy}
\label{sec:res-quality}

Before testing discovery, we check that the operators it starts from are accurate enough. The training data matter most. Replacing the low fidelity \textsc{Abl-2D} corpus with the high fidelity \nsD{} corpus reduces one step error from $0.155$ to $0.0005$, and at matched volume and resolution the generating solver alone accounts for about $200\times$ of this reduction, far more than resolution or training volume (Table~\ref{tab:ns}, Appendix~\ref{app:qualabl}).

The training curriculum matters less. In this curriculum the operator first learns to predict one step ahead, then gradually longer rollouts of its own predictions, up to 16 steps. It stabilizes 16 step rollouts on \textsc{Abl-2D} ($0.309$, against $0.965$ with one step training and $0.93$ for persistence, which copies the last frame forward) but is unnecessary on \nsD{}, where one step training is already stable ($0.0225$, against $0.038$ with the curriculum).

\label{sec:res-nsfinal}With high fidelity data, the 23M parameter NOS operator composes Navier--Stokes zero-shot at parity with the released NOS checkpoint at $128^2$ ($0.0036$ against $0.0045$) and reaches $0.0031\pm0.0004$ at $256^2$ over three seeds (Table~\ref{tab:nsresults}). Together, these results show that the supervised operators are accurate once trained on high fidelity data, so the errors reported in the following sections can be attributed to discovery rather than to the operators it starts from.

\subsection{Discovery in 1D}
\label{sec:res-ssl}

We first test whether unlabeled mixtures are sufficient to recover reusable mechanism operators. Table~\ref{tab:discovery} reports recovery, selection and utility for our procedure and its alternatives. Discovery observes only mixtures and uses the physical mechanism basis only for \textit{post-hoc} identification.


Exact assignment is important for recovery. The exact $w$ step recovers four out of five families in a single run and all five after pooling runs, whereas amortized assignment recovers at most two families per run (Appendix~\ref{app:amortized}). Operator representation also matters: the same discovery procedure identifies five out of five families with the codebook architecture, compared with three of five with the control architecture. The codebook model recovers the active mechanisms of held out mixtures at $95\%$ accuracy against $46\%$ permutation chance, and identification is unchanged with an independently implemented reference solver (Appendix~\ref{app:protocol:crosssolver}).

These choices remain label-free during discovery. Cluster count and run selection are based on held-out prediction on unlabeled data, while the physical identification metric is used only for \textit{post-hoc} evaluation (Appendix~\ref{app:opdist}). The discovered libraries can also be composed into simulators: their composed error stays within about twice that of the exact operators composed the same way, and most of the remaining gap comes from the nonlinear mechanisms (Figure~\ref{fig:panels}, appendix).


\begin{table}[t]
\centering\small
\caption{Discovery on \mixD{}, five hidden mechanisms. \emph{Recovery}: mechanisms identified out of five (\S\ref{sec:protocol}). \emph{Held-out selection}: accuracy of the active mechanisms predicted on unseen mixtures (permutation chance in parentheses). \emph{Utility}: composed rollout error on held-out mixtures (lower is better). --- marks a quantity the method does not produce. The full method uses the NOS architecture with its codebook; Table~\ref{tab:discovery-full} (appendix) reports both architectures, both gradings, and oracle utilities.}
\label{tab:discovery}
\resizebox{\linewidth}{!}{%
\begin{tabular}{lccc}
\toprule
Method & Recovery & Held-out selection & Utility \\
\midrule
amortized assignment (best of 4 formulations, Appendix~\ref{app:amortized}) & 0/5 & --- & 0.377 \\
SPLICE, single run (Eqs.~\ref{eq:wstep}--\ref{eq:phistep}) & 4/5 & 57.9\% $\pm$ 7.7 (51\%) & 0.147 \\
SPLICE, full (ensemble pooling, predictive $K$) & \textbf{5/5} & \textbf{95.0\%} (46\%) & 0.154 \\
\midrule
PDE-FIND (true symbolic terms given) \citep{rudy2017data} & 5/5 & --- & 0.056 \\
LEADS-style (environment ids given) \citep{yin2021leads} & 4/5 & --- & --- \\
\bottomrule
\end{tabular}}
\end{table}


\paragraph{Baselines.}
PDE FIND \citep{rudy2017data}, given the true symbolic terms, approaches oracle performance on clean data, while our operators require no symbolic prior. Under $10\%$ noise, the two approaches reach similar simulation utility (Appendix~\ref{app:noise}). LEADS \citep{yin2021leads} identifies four of five primitives even when environment identities are provided, compared with five of five for our exact fit procedure. 
Its shared component mixes advection and diffusion and misses dispersion, whereas our procedure produces separate, identifiable operators.

\subsection{Discovery in 2D: decomposing coupled Navier--Stokes dynamics}\label{sec:res-nsssl}\label{sec:res-2d}

We next test whether discovery transfers to 2D dynamics and, more importantly, whether it can separate mechanisms that are never observed in isolation. We use the NOS operator of \S\ref{sec:res-quality} and evaluate identification against independently specified per-family null hypotheses; Table~\ref{tab:nsresults} summarizes the results.

On a mixture of 128 Euler and 128 diffusion trajectories with identities hidden, the discovered library composes Navier--Stokes at essentially the same accuracy as the supervised library, $0.0038$ versus $0.0036$. At native $256^2$, the full discovery loop identifies both mechanisms on $10/10$ seeds and composes at $0.0028 \pm 0.0025$ (median $0.0016$), against $0.0031 \pm 0.0004$ (median $0.0033$) for the supervised library.

The more demanding test uses coupled Navier--Stokes trajectories, where Euler and diffusion act simultaneously at every step and no pure mechanism trajectories are available. Discovery recovers both mechanisms on all three seeds. 
At native $256^2$, discovery on coupled flows, warm started from the supervised library, identifies both mechanisms on all ten seeds and composes the coupled dynamics at $0.0116 \pm 0.0014$ (median $0.0118$).

The same adaptation applies to real data. On \texttt{shear\_flow} from The Well \citep{ohana2024well}, adapting the library on 32 unlabeled realizations, disjoint from the evaluated ones, makes its operators identify as Euler on three of three seeds, where the unadapted library missed, and lowers the composed transfer error (VRMSE) from $0.868$ to $0.818$ on the same base model, more than twice the seed spread, though short of the pre-registered target (Appendix~\ref{sec:res-transfer}).

\begin{figure}[t]
\centering
\begin{minipage}[c]{0.66\linewidth}
    \centering
    \includegraphics[width=\linewidth]{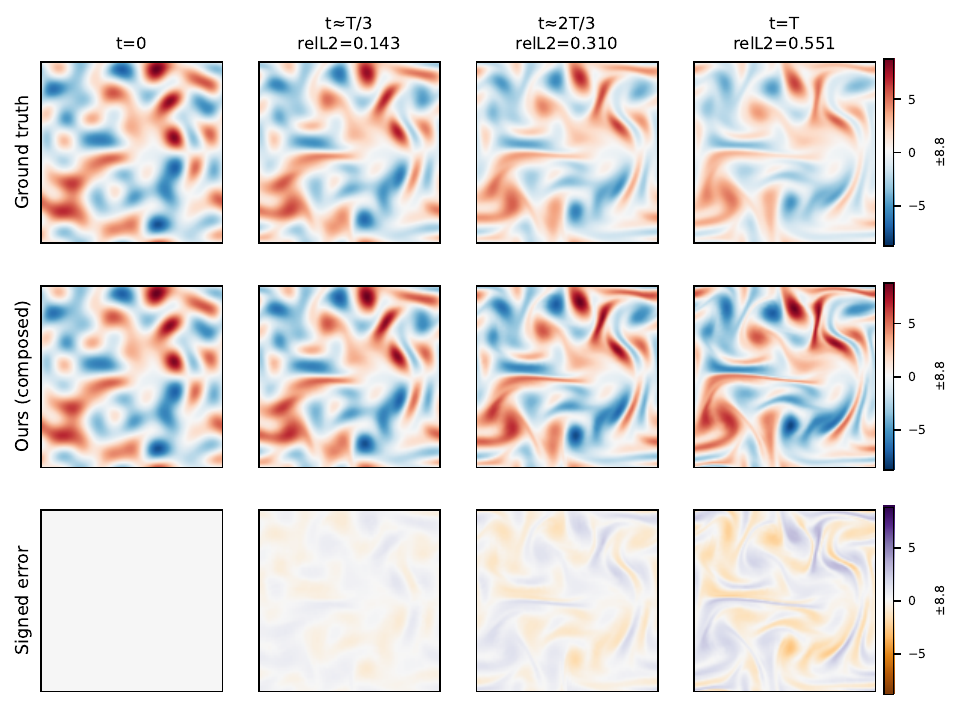}
\end{minipage}\hspace{0.01\linewidth}%
\begin{minipage}[c]{0.31\linewidth}
    \centering
    \includegraphics[width=\linewidth]{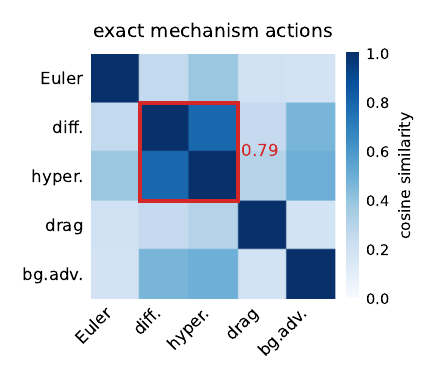}\\[2pt]
    \includegraphics[width=\linewidth]{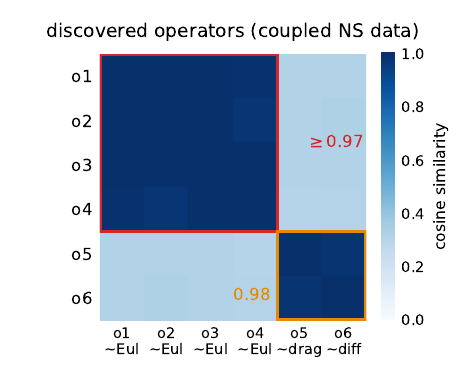}
\end{minipage}
\caption{\textbf{Zero-shot composition with a label-free library.}
\textbf{Left:} exact solution (top), rollout composed from discovered operators (middle) and their difference (bottom), at four times of a 60-step rollout on a coupled flow never seen in training ($128^2$). Reading each column from top to bottom, the positions and shapes of the vortices match throughout, and the error grows mainly in intensity. The true flow fades because viscosity and drag dissipate its energy; the composed rollout does not, because the beam selects only advection operators and each operator normalizes the amplitude of its input window, so this dissipation is not represented (Appendix~\ref{app:disc2d}).
\textbf{Right:} similarity between mechanism actions, for exact (top) and discovered (bottom) operators. Euler is nearly orthogonal to the overlapping smoothing mechanisms, hence identifiable (\S\ref{sec:res-negative}).}
\label{fig:nscomp}
\end{figure}

Figure~\ref{fig:nscomp} shows one composed rollout. The beam selects only the advection operators, although the mixture assignment attributes activity to both mechanisms: an operator can implement an identifiable mechanism without improving a particular composition (\S\ref{sec:res-negative}).

The same experiments show that discovery extends a pretrained library rather than reliably creating one from scratch. Random initialization collapses onto diffusion, while reseeding a low-weight operator recovers the missing Euler family on all three tested 2D seeds. We treat this as supporting evidence for library growth; detailed reseeding and leave-one-mechanism-out experiments are reported in Appendix~\ref{app:discovery-growth}.

Taken together, removing the labels loses nothing relative to the supervised NOS operators. The discovered library identifies Euler and diffusion, composes Navier--Stokes as accurately as its supervised counterpart and more accurately than the released NOS checkpoint, and recovers the same mechanisms from coupled flows, where no labeled data can exist.

\begin{table}[t]
\centering\small
\caption{Zero-shot Navier--Stokes composition (NRMSE on the released test trajectories, 10 steps, lower is better; persistence scores $0.012$ at $128^2$ and $0.017$ at $256^2$). \emph{Supervised}: trained on labeled pure-mechanism trajectories. \emph{Label-free}: discovered from the same trajectories with mechanism labels hidden, or from coupled flows where no pure trajectory exists; labels enter only post-hoc identification. Mean $\pm$ sd over seeds, median in parentheses. NOS is used composed, like our libraries; its direct rollout reaches $0.021$ at $128^2$ and $0.0016$ at $256^2$ (Appendix~\ref{app:nos}). Full comparison in Table~\ref{tab:nsfinal}.}
\label{tab:nsresults}
\resizebox{\linewidth}{!}{%
\begin{tabular}{lcccc}
\toprule
Training data & Released NOS & Supervised (ours) & Label-free (ours) & Identified \\
\midrule
pure mechanisms, $128^2$ & 0.0045 & $0.0036 \pm 0.0006$ & $0.0038 \pm 0.0001$ & all seeds \\
pure mechanisms, $256^2$ & 0.0056 & $0.0031 \pm 0.0004$ ($0.0033$) & $0.0028 \pm 0.0025$ ($0.0016$) & 10/10 \\
coupled flows, $256^2$ & n/a & n/a (no labeled data) & $0.0116 \pm 0.0014$ ($0.0118$) & 10/10 \\
\bottomrule
\end{tabular}}
\end{table}

\subsection{Growing a mechanism absent from pretraining}
\label{sec:res-reach}

Adaptation only moves codes inside a fixed network (\S\ref{sec:method-grow}), so it cannot reach a mechanism the network never learned. We therefore pretrain the network without Euler and ask whether Euler can be added afterwards, from trajectories that mix it with diffusion, without labels and without degrading what the network already knows.

Freezing the network and growing a new operator achieves both. The frozen library routes to the new operator the windows it reconstructs poorly, which isolates the Euler windows (recall $1.0$, precision $0.94$ to $0.95$), and the new operator is identified as Euler on all seeds. Its one step error matches retraining the whole network with replay of diffusion, at an eighth of the compute, and since no pretrained weight changes, diffusion is preserved exactly, whereas retraining degrades it (Table~\ref{tab:reach}). Over ten steps, replay remains more accurate, so growth trades some long horizon accuracy for preserving the library. Appendix~\ref{app:reach} reports labeled routing, retraining without replay, and rollout-trained variants; growth from coupled flows remains open.

\begin{table}[!t]
\centering\small
\caption{Adding Euler to a network pretrained without it, without labels. Errors are composed relative $L^2$ on held out Euler windows at rollout steps 1 and 10; diffusion error relative to the original library. Ranges over three seeds. Full comparison in Table~\ref{tab:reach-full}.}
\label{tab:reach}
\begin{tabular}{lcccc}
\toprule
Method & Step 1 & Step 10 & Diffusion & GPU h \\
\midrule
Frozen, no growth & $0.221$ & $0.88$ & $1.00\times$ & 0 \\
Retrain with replay & $[0.061, 0.073]$ & $[0.19, 0.22]$ & $[1.06, 1.42]\times$ & 9 \\
Frozen + new operator (ours) & $[0.057, 0.060]$ & $[0.32, 0.34]$ & $1.00\times$ & 1.5 \\
\bottomrule
\end{tabular}
\end{table}

\subsection{Limits}
\label{sec:res-negative}

The experiments also reveal four challenges for the approach. 

\emph{Mechanism separability.} Identification depends on different mechanisms producing sufficiently distinct actions. In 2D, diffusion, hyperdiffusion, and drag act almost identically, and drag aliases to diffusion in coupled discovery on all three tested seeds, whereas Euler, whose action is distinct, is recovered from coupled flows. Similar mechanisms therefore share one library slot (Appendix~\ref{app:protocol}).

\emph{Composability.} Physical identity does not guarantee that an operator improves a particular composition. On the flow of Figure~\ref{fig:nscomp}, forcing the identified diffusion operator into the composition increases the rollout error from $0.24$ to $0.68$; the effect depends on how the operator enters the composition, not on its accuracy as diffusion (Appendix~\ref{app:disc2d}).

\emph{Operator representation.} The operators standardize each input window by its mean and scale, so mechanism identity is recovered while absolute coefficient magnitude is not, which makes amplitude dependent mechanisms less natural for the current representation (Appendix~\ref{app:opdist}).

\emph{Splitting regime.} Finally, composition assumes that mechanisms can be applied one after the other with small error. In developed turbulence they interact too strongly for this: on turbulent flows from The Well, the beam finds no useful composition, and the rollout stays sharper than the true flow, which smooths out over time (Appendix~\ref{app:trl2d}). A learned operator for the interaction between mechanisms is a natural extension. Without fine-tuning, the 2D library composes \texttt{shear\_flow} stably where the beam selects diffusion, diverges where it selects Euler, and declines to compose \texttt{rayleigh\_benard}, whose buoyancy and walls lie outside the library (Appendix~\ref{sec:res-transfer}).

\section{Discussion and conclusion}
\label{sec:discussion}
\label{sec:conclusion}

We found that the operator splitting of physics can be learned from unlabeled trajectories and checked without labels. SPLICE adapts a library pretrained on synthetic single-mechanism data, where labels are free, to unlabeled data, and names the mechanism each operator implements, or reports that it cannot. Hidden mechanisms of 1D mixtures are recovered, the Euler and diffusion operators of Navier--Stokes are identified from coupled flows, and a missing mechanism grows back from its own residuals. A mechanism absent from pretraining can also be added as a new operator, without labels and without forgetting, at an fifth of the compute of retraining, though retraining remains more accurate over long rollouts. Learning without labels costs no accuracy and often improves it: at $256^2$, the adapted library composes more accurately than the supervised library it starts from on seven of ten seeds, halving its error on the median run, and more accurately than the released NOS model.



Because every component is identified, a failing simulation points to the operator responsible, and the remaining limits, such as mechanisms with nearly identical actions sharing one operator, can be measured directly (\S\ref{sec:res-negative}).




This suggests a way to build physics simulators: pretrain a library where labels are free, adapt it where they are not, and compose from identified parts. Next steps include growing new operators from coupled flows, composition-aware training, and other modalities, where labels are even scarcer. After pretraining, mechanism labels can simply be removed, and reference solvers suffice to tell what each operator learned.





\subsubsection*{Reproducibility statement}
Appendix~\ref{app:protocol} specifies the identification protocol in full; the hyperparameter tables and reproducibility notes are in the appendix, and every printed number traces to a logged artifact shipped with the supplementary material, alongside a pre-registration bundle of six records whose manifest states, for each, the criterion written before grading and whether it was met. Code at 
\url{https://github.com/LucasStill/SPLICE}

\subsubsection*{AI use statement}
Large language model assistants were used for drafting and editing text, writing experiment and analysis code, and orchestrating experiment runs. All quantitative claims trace to code; the authors checked every reported number against it and reviewed and take responsibility for all content.

\bibliographystyle{plainnat}
\bibliography{references}

\appendix
\section{Experimental Details}
\label{app:experimental}

\subsection{Notation}
\label{app:notation}

Table~\ref{tab:notation} collects the symbols used in the paper; dataset and library names are in Table~\ref{tab:data}.

\begin{table}[!htbp]
\centering\small
\caption{Symbols (top) and architectures with their instances (bottom).}
\label{tab:notation}
\resizebox{\linewidth}{!}{%
\begin{tabular}{ll}
\toprule
Symbol & Meaning \\
\midrule
$u_t \in \mathbb{R}^{C\times N}$ (1D), $\mathbb{R}^{C\times N\times N}$ (2D) & discretized field at step $t$; $C$ channels, $N$ grid points/axis \\
$u_{0:T}$ & a trajectory; $u_{t-L:t}$ a length-$L$ context window \\
$A_m$, $c_m \ge 0$ & mechanism $m$ (diffusion, advection, \dots) and its coefficient \\
$E$ & encoder: context window to latent code \\
$\theta = E(u_{t-L:t}) \in \mathbb{R}^d$ & latent code (selects mechanism and coefficient); $d$ its dimension \\
$f_\theta$ & conditioned vector field; continuous dynamics $\mathrm{d}v/\mathrm{d}s = f_\theta(v)$ \\
$\Phi_\theta(u_t) = \mathrm{RK4}[f_\theta](u_t)$ & one solver-operator step (Eq.~\ref{eq:step}) \\
$\Phi_A^{1/2}\!\circ\Phi_B\circ\Phi_A^{1/2}$ & one Strang-composed step; local error $O(\Delta t^3)$ \\
$\mathcal{L}=\{\Phi_{\theta_1},\dots,\Phi_{\theta_M}\}$ & operator library ($M$ mechanisms) \\
$\hat\Phi_k$, $\hat f_k$ & discovered operator $k$ and its vector field; $\hat\phi$ their codes \\
$w \in \mathbb{R}^K_{\ge 0}$ & per-trajectory mixture weights over discovered operators (w-step, Eq.~\ref{eq:wstep}) \\
$z_t$; $\Delta z_t = z_{t+1}-z_t$ & standardized frame at time $t$; finite-difference estimate of $\partial_t u$ \\
$\hat\Phi^{(h)}_{\hat\phi,w}$ & $h$-step rollout of the weight-mixed discovered field ($\phi$-step, Eq.~\ref{eq:phistep}) \\
$\lambda$ & weight of the total-weight penalty in the w-step \\
$K$ & number of discovered operators kept (cluster count for merging pooled runs) \\
$\mathcal{B}$; $\mathcal{S}_m$ & probe set of fixed synthetic fields; exact reference solver step for mechanism $m$ (App.~\ref{app:protocol}) \\
NRMSE & per-step-mean relative $L_2$ error over a 16-step rollout \\
\midrule
Architecture / instance & Meaning \\
\midrule
NOS architecture & \citet{serrano2026test} architecture (instantiated 1D and 2D) \\
control architecture & our independent second architecture (instantiated 1D and 2D) \\
NOS operator & NOS architecture without codebook and in-context conditioning (our reproduction) \\
full NOS method & NOS architecture with codebook, in-context, and substep ramp \\
2D instance & trained on \nsD{}; source of all NS numbers \\
1D instance & trained on \primD{}; source of all breadth numbers \\
\bottomrule
\end{tabular}}
\end{table}

\subsection{Datasets and mechanisms}
\label{app:data}

\begin{table}[!htbp]
\centering\small
\caption{Datasets (top) and operator libraries (bottom). ``Labels''$=$mechanism identity per trajectory.}
\label{tab:data}
\resizebox{\linewidth}{!}{%
\begin{tabular}{llll}
\toprule
Name & Content & Role & Labels \\
\midrule
\primD{} & 1D, 8 mechanisms $\times$ dense coefficients (spectral solver): advection, & SL training & yes \\
        & diffusion, dispersion, nonlinear advection, decay, hyperdiffusion, & & \\
        & cubic advection, bistable reaction & & \\
\mixD{} & 1D, hidden random 1--2-mechanism mixtures of 5 \primD{} mechanisms & SSL training & hidden (grading only) \\
        & (advection, diffusion, dispersion, nonlinear advection, decay) & & \\
\evalN{} & 1D, nine named PDEs (Table~\ref{tab:eqdefs}), each a 2- to 4-term & zero-shot test & --- \\
        & composition of \primD{} mechanisms with jittered coefficients & & \\
\nsD{} & 2D vorticity: pure Euler and pure diffusion training trajectories ($2048$ traj.) & 2D SL training / test & yes / --- \\
       & + NS test set ($3072$ traj.); public release of \citealp{serrano2026test} & & \\
\textsc{Abl-2D} & 2D vorticity, self-generated, lower-fidelity solver ($256$ trajectories) & \emph{data ablation only} (\S\ref{sec:res-quality}) & yes \\
\midrule
\slib{2} & supervised library, 2 mechanisms (NOS scale) & baseline & yes \\
\slib{8} & supervised library, 8 mechanisms (trained on \primD{}) & ours & yes \\
\dlib{} & discovered library (trained on \mixD{} or unlabeled \nsD{} mixtures) & ours & none \\
\bottomrule
\end{tabular}}
\end{table}

Each \primD{} mechanism is one differential operator acting on the field $u$: \emph{advection} $-c\,\partial_x u$ transports the field at speed $c$; \emph{diffusion} $\nu\,\partial_{xx}u$ smooths it; \emph{dispersion} $-\partial_{xxx}u$ makes phase speed depend on wavenumber; \emph{nonlinear self-advection} $-u\,\partial_x u$ and \emph{cubic advection} $-u^2\,\partial_x u$ steepen the field toward shocks; \emph{decay} $-\gamma u$ damps it uniformly; \emph{hyperdiffusion} $-\nu_4\,\partial_{xxxx}u$ damps high wavenumbers more sharply than diffusion; and \emph{reaction} $\rho\,u(1-u)$ (bistable form $\rho\,u(1-u^2)$) grows or saturates the field pointwise. In 2D the \emph{Euler} operator advects vorticity by its own induced velocity. Table~\ref{tab:eqdefs} gives the exact form of each \evalN{} equation.

\begin{table}[!htbp]
\centering\small
\caption{The nine \evalN{} equations, each a composition of \primD{} mechanisms. Free coefficients are jittered within the ranges of Table~\ref{tab:hparams}; the dispersion coefficient is fixed to unity. Signs and the exact reaction terms follow the data generators.}
\label{tab:eqdefs}
\resizebox{\linewidth}{!}{%
\begin{tabular}{lll}
\toprule
Equation & PDE & Active mechanisms \\
\midrule
Burgers & $\partial_t u = -u\,\partial_x u + \nu\,\partial_{xx}u$ & nonlinear advection, diffusion \\
KdV & $\partial_t u = -u\,\partial_x u - \partial_{xxx}u$ & nonlinear advection, dispersion \\
mKdV & $\partial_t u = -u^2\,\partial_x u - \partial_{xxx}u$ & cubic advection, dispersion \\
Allen--Cahn & $\partial_t u = \nu\,\partial_{xx}u + \rho\,u(1-u^2)$ & diffusion, bistable reaction \\
advection+reaction & $\partial_t u = -c\,\partial_x u + \rho\,u(1-u)$ & advection, reaction \\
damped KdV & $\partial_t u = -u\,\partial_x u - \partial_{xxx}u - \gamma u$ & nonlinear advection, dispersion, decay \\
hyperviscous KdV & $\partial_t u = -u\,\partial_x u - \partial_{xxx}u - \nu_4\,\partial_{xxxx}u$ & nonlinear advection, dispersion, hyperdiffusion \\
KdV--Burgers--decay & $\partial_t u = -u\,\partial_x u + \nu\,\partial_{xx}u - \partial_{xxx}u - \gamma u$ & nonlin.\ advection, diffusion, dispersion, decay \\
adv.--diff.--decay--hyperdiff. & $\partial_t u = -c\,\partial_x u + \nu\,\partial_{xx}u - \gamma u - \nu_4\,\partial_{xxxx}u$ & advection, diffusion, decay, hyperdiffusion \\
\bottomrule
\end{tabular}}
\end{table}

\subsection{Training configurations and architecture}
\label{app:training}\label{app:implementation}

Table~\ref{tab:hparams} lists the generation and training settings; the shared architecture is shown in Figure~\ref{fig:arch}.

\begin{table}[!htbp]
\centering\small
\caption{Data generation and training configurations. 1D: spectral solver, five discovery mechanisms shown (the full \primD{} corpus has eight). 2D: pseudo-spectral vorticity solver; data generated at $256^2$ and trained at $128^2$. Parameter counts measured from checkpoints.}
\label{tab:hparams}
\resizebox{\linewidth}{!}{%
\begin{tabular}{lll}
\toprule
 & Setting & Value \\
\midrule
1D & grid, steps per trajectory & $n_x{=}256$, $100$ steps ($L{=}16$, $T{=}4$, $\Delta t{=}0.04$) \\
 & coefficient ranges & advection $[0.1,2.0]$, diffusion $[0.01,0.4]$, dispersion $[0.1,1.0]$, \\
 & & nonlinear advection $[0.1,0.5]$, decay $[0.1,1.0]$ \\
 & training & $10^5$ iterations, batch $32$, lr $3{\times}10^{-4}$ (cosine), weight decay $10^{-4}$ \\
 & model & $16.30$M parameters \\
\midrule
2D & grid, snapshots & $128^2$ (generated at $256^2$), $50$ snapshots ($T{=}4$, $\Delta t{=}4/250$) \\
 & coefficient ranges & diffusion $\nu \in [10^{-3},10^{-2}]$, hyperdiffusion $\nu_4 \in [10^{-4},10^{-3}]$, \\
 & & drag $[0.1,0.6]$, background advection $U \in [0.5,1.5]$, Euler (no free coeff.) \\
 & training & $4{\times}10^4$ iterations, batch $8$, lr $3{\times}10^{-4}$ (warmup $10^3$), \\
 & & $4$ integration substeps, rollout-horizon ramp $1 \to 16$ (parity); $1 \to 8$ (breadth-suite operator) \\
 & corpus & $1024$ Euler ${+}$ $1024$ diffusion trajectories; released NS test set (native $256^2$) \\
 & model & $23.61$M parameters \\
\bottomrule
\end{tabular}}
\end{table}

\subsection{Evaluation settings and per-equation results}
\label{app:evaluation}\label{app:pereq}\label{app:pereq-figanchor}

\begin{figure}[!htbp]
\centering
\fbox{\includegraphics[width=0.48\linewidth]{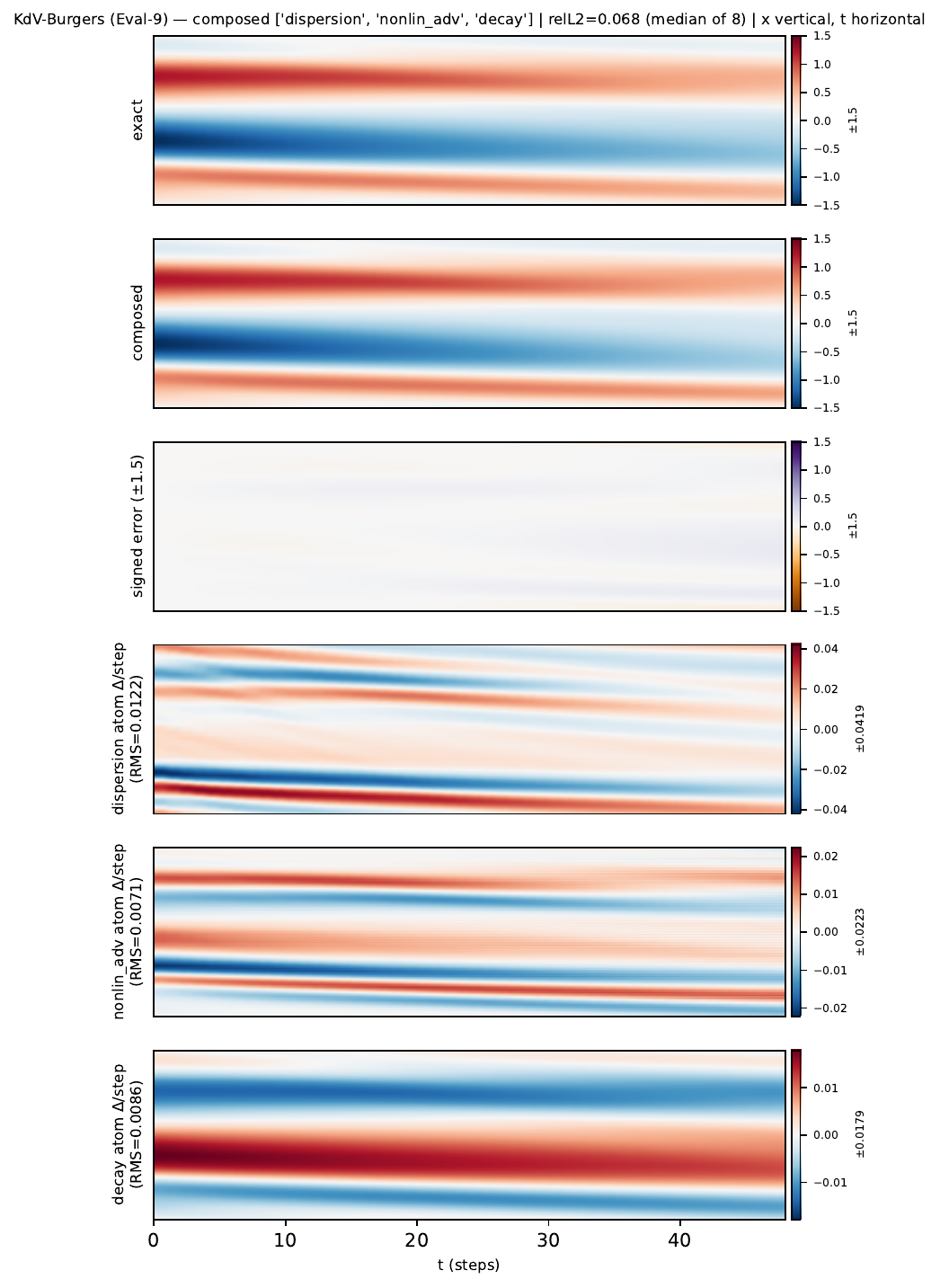}}
\caption{Zero-shot composition of the four-term \evalN{} equation KdV-Burgers with decay (nonlinear advection, dispersion, diffusion, decay) by a five-family supervised library (space-time; the upper-median of 8 test trajectories by composed rollout error, relative $L_2$ $0.068$, set median $0.065$). Rows: exact solution; composed rollout; signed error; per-operator increments on their own scales. Compositions are capped at three operators, so one of the four terms must be dropped; the beam keeps dispersion, nonlinear advection, and decay and drops diffusion, a dissipative-for-dissipative substitution consistent with the separability limits of \S\ref{sec:res-negative}. The nonlinear operator carries the steepening.}
\label{fig:1dcomp}
\end{figure}

\begin{figure}[!htbp]
\centering
\fbox{\includegraphics[width=0.72\linewidth]{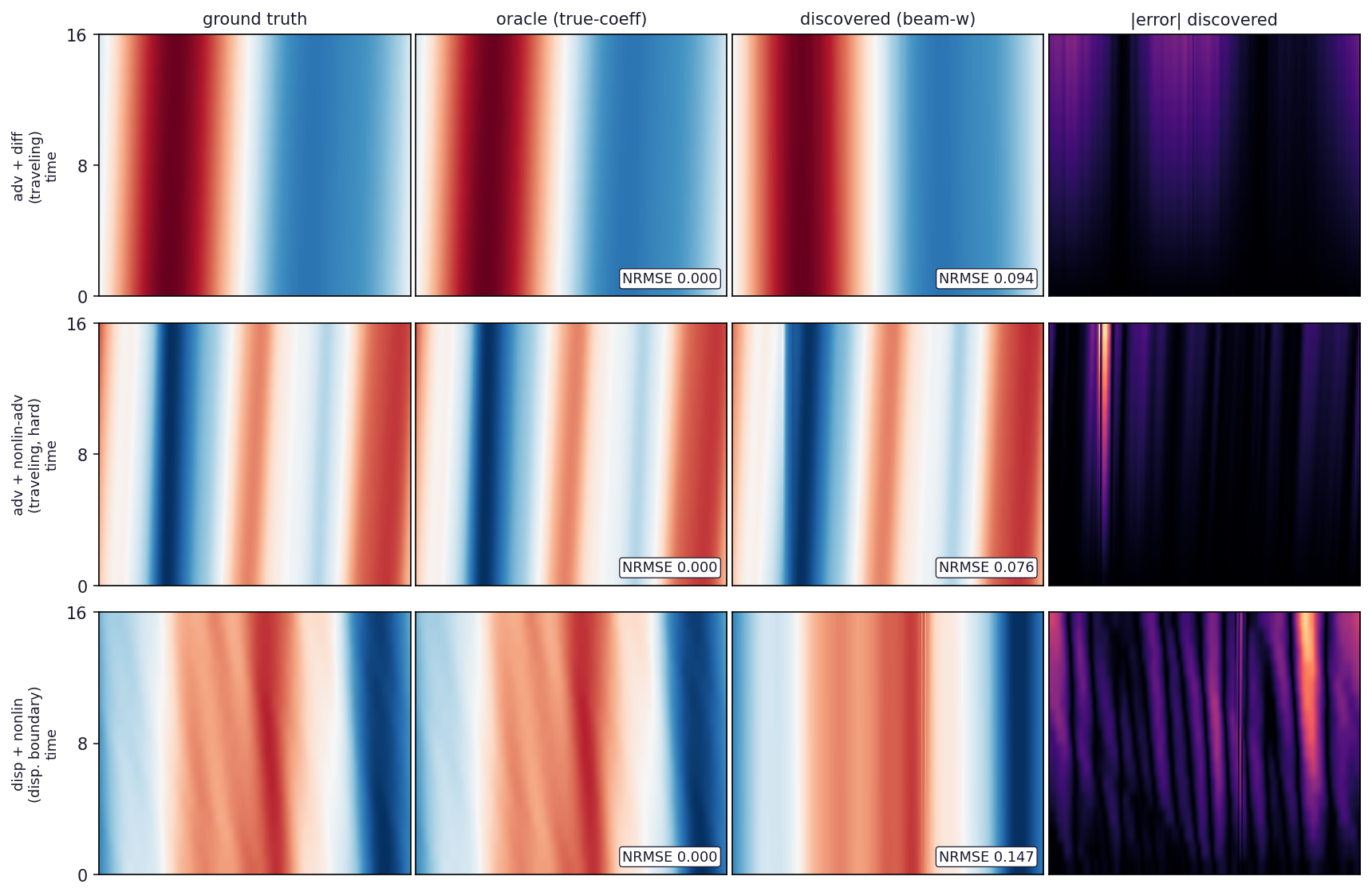}}
\caption{Rollouts on three held-out 1D mixtures (\S\ref{sec:res-ssl}): ground truth; oracle operators; discovered library \dlib{}; absolute error (shared scale). Top: advection${+}$diffusion, cleanly reproduced ($0.094$). Middle: a traveling advection${+}$nonlinear case, where the discovered library reproduces the transport ($0.076$). Bottom: an advection${+}$dispersion case at the dispersion-limited boundary ($0.147$): the discovered rollout cannot travel because the residual transport there is dispersion-driven and dispersion is the rarest mechanism in the discovered pool (Appendix~\ref{app:opdist}).}
\label{fig:panels}
\end{figure}

Table~\ref{tab:pereq} breaks the \evalN{} mean of Table~\ref{tab:battery} down per equation for the NOS-operator \slib{8} library (three seeds, re-measured; Appendix~\ref{app:repro}). The error is spread across equations ($0.048$ to $0.123$), and the four-term systems ($0.107$, $0.067$) are no harder than the two-term ones. Figure~\ref{fig:1dcomp} shows one composed example and Figure~\ref{fig:panels} three held-out discovery rollouts.

\begin{table}[!htbp]
\centering\small
\caption{Per-equation composed NRMSE on \evalN{} (NOS operator, \slib{8}; mean over 3 seeds, re-measured 2026-09, Appendix~\ref{app:repro}).}
\label{tab:pereq}
\begin{tabular}{lc@{\qquad}lc}
\toprule
Equation & NRMSE & Equation & NRMSE \\
\midrule
Burgers & 0.091 & damped KdV & 0.112 \\
KdV & 0.123 & hyperviscous KdV & 0.123 \\
mKdV & 0.048 & KdV-Burgers-decay (4-term) & 0.107 \\
Allen-Cahn & 0.074 & adv.-diff.-decay-hyperdiff.\ (4-term) & 0.067 \\
advection+reaction & 0.115 & \textbf{mean} & \textbf{0.096} \\
\bottomrule
\end{tabular}
\end{table}

\subsection{Supervised operator fidelity ablations}\label{app:qualabl}

Table~\ref{tab:ns} collects the Navier--Stokes ablations behind \S\ref{sec:res-quality}. As a reading aid for rollout NRMSE: below $0.1$ is usable, above $0.3$ is not. A monolithic FNO trained with the curriculum on the same $2048$ pure-mechanism trajectories does not beat persistence on the released \nsD{} slice ($0.031$ over 16 steps, best of a sweep) and ties our operator on the dynamic \textsc{Abl-2D} slice ($0.31$ each); we therefore make no 2D claim of composition over monoliths. In the one-step against curriculum comparison, batch sizes are not perfectly matched, because the 16-frame curriculum does not fit in memory at $256^2$.

\begin{table}[!htbp]
\centering\small
\caption{Navier--Stokes quality ablations (per-step-mean NRMSE, $128^2$). \textsc{Abl-2D} rows: 16-step rollouts, persistence $0.93$. \nsD{} data-control rows: released test trajectories over the 10-step released horizon (persistence $0.012$); the \nsD{} pushforward rows in the objective control are 16-step rollouts on the same trajectories. Left: training-objective control on \textsc{Abl-2D} and, below the rule, on \nsD{} (both objectives from scratch with the pushforward trick). Right: data control and, below the rule, its split into volume, resolution, and solver at a shorter matched budget (hence the $0.0031$ anchor in that block). All models supervised.}
\label{tab:ns}
\resizebox{\linewidth}{!}{%
\begin{tabular}{lc@{\qquad}lcc}
\toprule
Objective control & rollout & Data control & rollout & one-step \\
\midrule
\textsc{Abl-2D}: persistence (floor) & 0.93 & small self-generated corpus (\textsc{Abl-2D}) & 0.31 & 0.155 \\
\textsc{Abl-2D}: one-step training & 0.965 (diverges) & NOS corpus (\nsD{}) & \textbf{0.0034} & 0.0005 \\
\textsc{Abl-2D}: + best post-hoc stabilizer & 0.267 & NOS model, same protocol & 0.021 & --- \\
\textsc{Abl-2D}: rollout curriculum & 0.309 (stable) & \emph{disentanglement (matched budget):} & & \\
\midrule
\nsD{}: one-step training (pushforward) & 0.0225 (stable) & \nsD{}, full volume, $128^2$ & --- & 0.0031 \\
\nsD{}: rollout curriculum (pushforward) & 0.038 & \nsD{}, 256 traj.\ (volume matched), $128^2$ & --- & 0.0007 \\
 & & \nsD{}, full volume, $64^2$ (resolution) & --- & 0.0071 \\
 & & \textsc{Abl-2D} solver, 256 traj., $128^2$ (solver) & --- & 0.155 \\
\bottomrule
\end{tabular}}
\end{table}

\paragraph{Stabilizers and corpus levers.} On \textsc{Abl-2D}, the best post-hoc stabilizer of one-step-trained operators reaches $0.267$ (other variants $0.341$ to $0.434$), none reaching the usability bar. At fixed architecture and recipe, the generating solver moves one-step error about $200\times$, volume contributes nothing measurable down to 256 trajectories, and resolution about $2\times$ per octave (Table~\ref{tab:ns}).

\paragraph{1D head-to-head.} The \evalN{} comparison of Table~\ref{tab:battery} uses one shared beam harness with an identical search budget for both architectures, with a cap on per-context search so that neither is advantaged by longer test-time compute. Across seeds, the identification distance of exact-fit discovered operators varies about $50\times$ less than that of amortized-selection discovered operators.

\section{Physical Identity Evaluation Protocol}
\label{app:protocol}

This appendix details criterion (i) of \S\ref{sec:protocol}: the probe set, the identification distance, its margin, the per-mechanism null threshold, and the absolute cap.

\subsection{Probe set, distance, and thresholds}
\label{app:protocol:probes}\label{app:protocol:fingerprint}\label{app:protocol:null}

\paragraph{Probe set.} All identification uses one fixed set of synthetic input fields, the probe set $\mathcal{B}$, built once, deterministically, and independent of any training data, trajectory, or label: (i) pure Fourier modes $\cos(kx),\sin(kx)$ over a fixed range of wavenumbers; (ii) localized Gaussian bumps at fixed widths and positions; (iii) random band-limited fields from a fixed seed.

\paragraph{Identification distance.} A discovered operator $\hat\Phi$ is graded by Eq.~\eqref{eq:fingerprint}: its one-step displacement $\hat\Phi(p)-p$ on each probe, against the displacement of each exact solver under the same integrator and time scale, with the probe interval pinned to each operator's $\theta$-implied step. Normalizing by the displacement rather than the full response removes the input state, which otherwise dominates the distance for small-step operators. A discovered operator is assigned to $A_m$ only if $m$ is its closest mechanism by a margin (the ratio of second-closest to closest distance; recovery is unchanged for ratios $1.0$ to $2.0$) and $d_m(\hat\Phi)$ falls below that mechanism's null threshold.

\paragraph{2D grading.} In 2D, each discovery seed is graded on its own probe set of three random band-limited vorticity fields (25 Fourier modes) drawn from a seed-specific generator. The distance compares four-frame rollouts from each probe with those of the exact solver, normalized by the root-mean-square of the reference rollout, and takes the minimum over eight candidate time scales (and, for diffusion, over a grid of viscosities). The null is the 5th percentile of the same distance over 24 random codes, drawn at the scale used to initialize discovered codes and passed through the seed's own trained network, so both sides use the same estimator and the same network.

\paragraph{Null calibration.} Thresholds are set per mechanism from a null of $d_m$ values produced by trained discovered operators identified with the other mechanisms, so the null represents a trained operator of the wrong kind, not noise; a discovered operator identifies as $A_m$ only below the 5th percentile of that null. Per-mechanism thresholds are needed because mechanisms differ widely in natural displacement size. An absolute cap ($d_m < 1.5$) also rejects diverged discovered operators that would otherwise identify by being the closest of uniformly poor fits.

\paragraph{Held-out selection.} For each held-out trajectory the w-step (Eq.~\ref{eq:wstep}) yields a support pattern over discovered operators, compared to the true active set. Chance comes from a permutation null (shuffled trajectory-label pairings, $10^3$ permutations); Table~\ref{tab:discovery} reports the permutation mean in parentheses.

\paragraph{Pre-registration and seeds.} Every model-selection criterion was committed in writing before its result was read; \S\ref{sec:res-ssl} reports both the criteria that held and those that failed (eigengap, silhouette). Headline recovery claims need two of three seeds.

\paragraph{Independent solver check.}\label{app:protocol:crosssolver}
Repeating identification with an independently implemented finite-difference reference solver in place of the spectral one leaves operator identity, identification outcomes, and codebook recovery unchanged ($5/5$ in each case), with a maximum change in self-distance of $0.18$ against a null distance near $1.0$.

\subsection{Original and discriminative grading; two architectures}
\label{app:protocol:audit}\label{app:twobackbone}

An earlier version of the distance normalized by the full response $\|\mathcal{S}_m(p)\|_2$ instead of the displacement. For small-step operators the probe $p$ then dominates, so near-identity operators score about $0.01$ against every mechanism and cannot be told apart. We therefore grade by the displacement, which isolates what an operator does rather than the state it receives; this discriminative distance (Eq.~\eqref{eq:fingerprint}) is used for all results. Because this correction followed the first results, the ``orig.'' column of Table~\ref{tab:discovery-full} keeps the original grading. The correction does not favor our method by construction: it lowers the control architecture from $5/5$ to $3/5$, held-out selection, which does not use the distance, agrees with it ($95\%$ against $46\%$ chance), and every later pre-registered experiment uses it.

The two gradings order the two architectures differently (Table~\ref{tab:discovery-full}). Under the original grading, pooled discovery identifies all five mechanisms with the control architecture and only decay with the NOS codebook architecture, whose small per-step displacements fall inside the null. Under the discriminative grading the codebook architecture identifies $5/5$ and the control architecture $3/5$, and held-out selection agrees ($95.0\%$ against $46\%$ chance, and $72.5\%$ against $61\%$). The control architecture's two misses are systematic (discovered diffusion operators assigned to decay, nonlinear-advection operators to dispersion) and persist for margin ratios $1.0$ to $2.0$; a stricter margin lowers both counts by one (codebook $4/5$, control $2/5$) without changing the ordering. Accuracy is comparable across the two: per-mechanism one-step error has median $0.010$ for both 1D NOS-architecture operators, and the retrained full NOS composes NS at $0.0053$ (Table~\ref{tab:nsfinal}).

\subsection{Identifiability and separability}
\label{app:ident}

\paragraph{Drop-one identifiability.} Before asking a selection rule to recover a mechanism, we ask whether the data can reveal it: we roll the exact solver with one true mechanism removed and report the NRMSE of that rollout against the full equation. A mechanism is identifiable at a horizon if this change exceeds three times the identified operator's rollout error. On a five-equation 2D suite (Euler advection, diffusion, hyperdiffusion, drag, and advection by a fixed background field; identifiable if the change is $\ge 0.15$), every advective mechanism is identifiable (drop-one change $0.71$ to $1.15$), while diffusion is identifiable in three of the equations only after raising its viscosity ($0.07$ to $0.14$ before, $0.19$ to $0.27$ after). On coupled NS, dropping Euler changes the 16-step rollout by $0.88$. On \evalN{}, $18$ of $21$ active mechanisms are identifiable at 16 steps (hyperdiffusion and reaction are not audited); the three below threshold are weak short-horizon terms (diffusion in KdV and mKdV, decay in damped KdV), so the $9/9$ mechanism-set result of Table~\ref{tab:battery} rests on the dominant terms.

\paragraph{Composing the 2D suite.} None of the five suite equations composes to the NRMSE $<0.1$ bar at $128^2$ ($0.32$ to $0.54$, against persistence $0.57$ to $1.14$). Standalone, supervised diffusion ($0.061$) and hyperdiffusion ($0.092$) pass, drag is borderline ($0.118$), and Euler ($0.62$; $0.43$ when trained on \nsD{}) and background advection ($0.25$) fail: pure advective transport compounds phase error over 16 steps at this scale. A pre-registered dissipative-only system (diffusion + hyperdiffusion + drag) also fails ($0.358$, persistence $0.381$), with diffusion never selected.

\paragraph{Aliasing.} With Euler, diffusion, and drag all active in coupled discovery, Euler and diffusion identify on $3/3$ seeds and drag on $0/3$, its discovered operators graded as diffusion. Horizon-resolved identification at $H \in \{4,8,16,32\}$ changes no supervised operator's mechanism: hyperdiffusion and drag are graded diffusion throughout. Across a viscosity sweep of the coupled pair ($\nu \in \{2,5,9\}\times 10^{-3}$), Euler and diffusion identify at every $\nu$ (margins $+0.043$ and $+0.015$). A fresh joint retraining (three seeds) replicates the aliasing. The dissipative side offers one identifiable slot, and the advective side identifies only through coupling.

\paragraph{Scope of the architectures.} Both architectures standardize each input window by its mean and standard deviation before the vector field acts, so a fixed discovered operator represents an operator only up to the window's mean and scale. Shift- and scale-equivariant differential operators (advection, diffusion, hyperdiffusion, dispersion, Euler) are represented exactly, and linear decay is exact on zero-mean fields. Amplitude-dependent reaction terms such as $\rho\,u(1-u)$ are represented only through the per-window code, which is why coefficient magnitude is not recovered (Appendix~\ref{app:opdist}).

\section{Discovery and Growth Ablations}
\label{app:ablations}\label{app:discovery-growth}

\subsection{Design of the discovery procedure}\label{app:design}

\paragraph{Exact versus amortized assignment.}\label{app:amortized}
We compared the exact w-step of Eq.~\eqref{eq:wstep} with four amortized alternatives. Three are selection heads over a frozen dictionary (softmax, Gumbel-softmax, sparsemax); their recovery equals the dictionary's by construction, and the best head reaches $40.3\%$ held-out selection against $44.5\%$ (deduplicated) and $39.5\%$ (single run) for the exact w-step, at a permutation chance of $20.9\%$. The fourth, an end-to-end encoder that predicts mixture weights and learns the discovered operators jointly, plateaus at two of five recovered mechanisms: selector error and operator error compensate each other, the loss decreases, and no discovered operator becomes sharp under the identification distance. Replacing amortized inference with the exact w-step lifts recovery to four of five. (The joint-encoder checkpoints predate the discriminative distance, so Table~\ref{tab:discovery-full} reports its recovery under the original grading only.)

\begin{table}[!htbp]
\centering\small
\caption{Complete discovery table on \mixD{} (5 hidden families), expanding Table~\ref{tab:discovery}: recovery under both the original (orig.) and the discriminative grading (Appendix~\ref{app:twobackbone}), held-out selection accuracy on unseen mixtures (permutation chance in parentheses), and utility through the shared composition harness against each architecture's oracle. The LEADS rows use a four-environment variant built for LEADS (identities given to LEADS only); --- marks cells undefined for a method's output and n.m.\ cells not measured, and the amortized row's selection is omitted as incomparable (fixed top-2 versus variable-cardinality sets).}
\label{tab:discovery-full}
\resizebox{\linewidth}{!}{%
\begin{tabular}{llcccccl}
\toprule
Method & Architecture & \multicolumn{2}{c}{Recovery} & Held-out & Utility & Oracle & Labels \\
 & & orig. & discr. & selection & & & \\
\midrule
amortized (best of 4 formulations, Appendix~\ref{app:amortized}) & control & 2/5 & 0/5 & --- & 0.377 & 0.063 & none \\
exact-fit (Eqs.~\ref{eq:wstep}--\ref{eq:phistep}), single run & control & 4/5 & 4/5 & 57.9\% $\pm$ 7.7 (51\%) & 0.147 & 0.063 & none \\
+ ensemble pooling + predictive $K$ & control & 5/5 & 3/5 & 72.5\% (61\%) & 0.124 & 0.063 & none \\
+ ensemble pooling + predictive $K$ & NOS + codebook & 1/5 & \textbf{5/5} & \textbf{95.0\%} (46\%) & 0.154 & 0.077 & none \\
\midrule
PDE-FIND (true terms in candidate basis) & --- & 5/5 & 5/5 & --- & 0.056 & --- & symbolic prior \\
LEADS-style shared field + per-environment residual \citep{yin2021leads} & control & --- & 4/5 & --- & --- & --- & environment ids \\
exact-fit, same four-environment variant & control & n.m. & 5/5 & n.m. & n.m. & n.m. & none \\
\midrule
\multicolumn{3}{l}{\emph{Where each construction can run}} & \multicolumn{2}{l}{\emph{supervised}} & \multicolumn{3}{l}{\emph{label-free}} \\
\multicolumn{3}{l}{NS composed zero-shot, $128^2$} & \multicolumn{2}{l}{$0.0036 \pm 0.0006$} & \multicolumn{3}{l}{$0.0038 \pm 0.0001$ (parity)} \\
\multicolumn{3}{l}{NS composed, native $256^2$} & \multicolumn{2}{l}{$0.0031 \pm 0.0004$ (median $0.0033$)} & \multicolumn{3}{l}{$0.0028 \pm 0.0025$ (median $0.0016$), $10/10$ identified} \\
\multicolumn{3}{l}{operators from coupled flows (\S\ref{sec:res-nsssl})} & \multicolumn{2}{l}{n/a: no labeled data exists} & \multicolumn{3}{l}{$10/10$ identified; $0.0116$ (native $256^2$)} \\
\multicolumn{3}{l}{regrowing a missing mechanism (\S\ref{sec:res-nsssl})} & \multicolumn{2}{l}{n/a} & \multicolumn{3}{l}{$3/3$ (supporting)} \\
\multicolumn{3}{l}{real data, \texttt{shear\_flow} adaptation (\S\ref{sec:res-transfer})} & \multicolumn{2}{l}{n/a} & \multicolumn{3}{l}{Euler identified $3/3$; $0.868 \to 0.818$} \\
\bottomrule
\end{tabular}}
\end{table}

\paragraph{Warm start versus random initialization.}\label{app:warmstart}
On the 2D mixture of \S\ref{sec:res-nsssl} (128 pure-Euler and 128 pure-diffusion trajectories, identities hidden), warm-started discovery recovers both mechanisms on seeds 0 and 1 (Euler fit $0.012$ to $0.028$), while random initialization recovers $1/2$ and $0/2$: its operators collapse onto diffusion and Euler is never recovered.

\paragraph{Fixed codes versus per-window refitting.}\label{app:refitting}\label{app:refit}
Per-window code refitting, the in-context alternative to composing with identified codes locked, does not help: $0.634$ refitted against $0.616$ locked on the dynamic slice.

\subsection{Library growth by reseeding}\label{app:reseed}

A discovered operator is dead when its held-out weight stays near zero (a duplicate, or near the identity), and the $\phi$-step never moves an operator that no trajectory selects. Reseeding re-initializes it by encoding a high-residual window, the slice of unlabeled data the current library reconstructs worst under Eq.~\eqref{eq:wstep}, so the new operator is planted where the error lives. This is the empty-cluster restart of $k$-means and the dead-element replacement of dictionary learning \citep{aharon2006k,mairal2010online}, transposed to operators. Keeping the best of $R$ candidates from the $R$ highest-residual windows removes the dependence on a single window; in \S\ref{sec:res-nsssl}, a missing mechanism never emerges without reseeding and emerges on three of three seeds with best-of-$R$. Dispersion is the exception: a pre-registered variant that scores windows by the odd-derivative fraction of the residual fails on three of three seeds, as does the total-residual control, with fresh discovered dispersion operators diverging. The bottleneck is training a stable discovered dispersion operator, not selecting windows.

\paragraph{Leave-one-family-out regrowth (negative).} We also tested regrowth when the backbone has never seen the target family: a backbone pretrained from scratch with the Euler family excluded, followed by the same pre-registered best-of-3 regrowth. On seed~0 of the three registered seeds, regrowth recovered diffusion only; no regrown operator was identified as Euler. The remaining two seeds each need a new backbone pretraining and are not reported. A pre-registered follow-up places the cause in the backbone rather than in discovery: on the same Euler-excluded network, a single code fitted directly on pure Euler windows, with labels and discovery's optimization budget, still identifies as diffusion ($d_{\mathrm{Euler}} = 0.050$ against an Euler null of $0.038$; nearest mechanism diffusion at $0.033$). Since discovery adapts operator codes within the pretrained network, it reaches only dynamics that network already represents. Adding such a mechanism requires changing or extending the network; Appendix~\ref{app:reach} and \S\ref{sec:res-reach} test both.

\subsection{The discovered operators in 1D and 2D}\label{app:opdist-group}

\begin{figure}[!htbp]
\centering
\fbox{\includegraphics[width=0.5\linewidth]{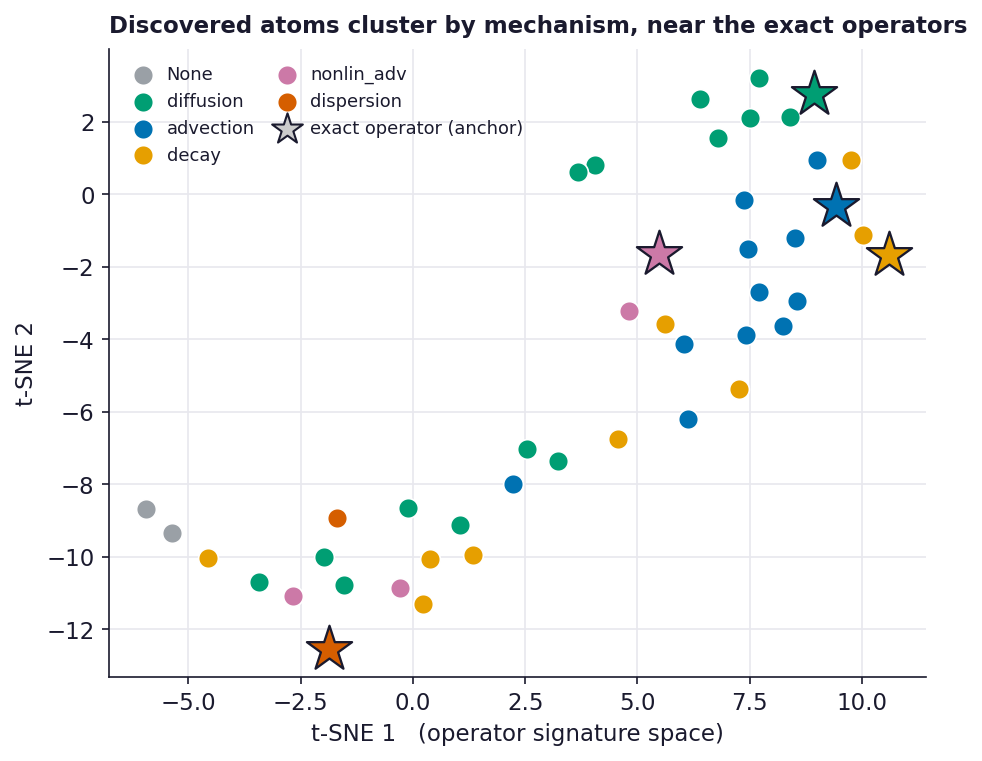}}
\caption{Discovered operators in operator-signature space (t-SNE) for the five-mechanism \mixD{} library, colored by graded mechanism; stars are the exact hand-coded operators. Discovered operators cluster near their mechanism's operator; all five mechanisms are recovered, with dispersion the rarest (a single discovered operator).}
\label{fig:opdist}
\end{figure}

\paragraph{Distribution and coefficient recovery.}\label{app:opdist}
Figure~\ref{fig:opdist} shows the discovered five-mechanism \mixD{} library in operator-signature space (a t-SNE of each discovered operator's action on the probe set): the discovered operators cluster by mechanism around the exact operators. The pool is uneven: 14 diffusion and 11 advection operators against a single dispersion operator. Dispersion is never learned on the harder eight-mechanism basis (\S\ref{sec:res-negative}). Coefficients are recovered far less reliably than identity: the deployed assignment weight correlates with the true mixture coefficient only for advection ($r{=}0.82$) and is near zero elsewhere (pooled $r{=}0.17$). Three analysis-level corrections raise the pooled correlation only to $0.275$, and a coefficient-equivariant training variant in the spirit of \citet{garrido2023self} does not help. The cause is not mixture identifiability: on held-out pure single-mechanism trajectories the weight again tracks the coefficient only for advection ($r{=}0.93$), because the operator standardizes each window by mean and scale (Appendix~\ref{app:ident}).

\paragraph{Cluster-count selection.}\label{app:kselection} Structure-based criteria for the cluster count $K$ used to merge pooled duplicates (eigengap, silhouette; pre-registered) select $K{=}4$ and recover $2/5$ under the original grading; held-out predictive utility selects $K{=}15$, inside the stable $5/5$ region. Under discriminative grading the pooled library identifies $3/5$ at every $K$, so $K$ governs redundancy rather than coverage.

\paragraph{Identified is not composable.}\label{app:disc2d}
On the dynamic coupled flow of Figure~\ref{fig:nscomp}, the composed library scores $0.243$ (rollout mean, 60 steps), and dropping diffusion from the exact solver changes the rollout by $0.098$, so diffusion matters. Yet forcing the identified diffusion operator into the composition degrades the rollout from $0.243$ to $0.675$, although its per-step increment behaves as diffusion (correlation $+0.955$ with $\nabla^2\omega$, $92\%$ of its energy below wavenumber $8$). The break depends on how the operator enters the composition: substituting it for an advection operator breaks the rollout ($0.2431 \to 0.6750$), while appending it as a third substep is harmless ($0.2413$). Amplitude does not explain it: amplitude-aware operators degrade identically under substitution ($0.2377 \to 0.6712$), and the correct and forced rollouts carry the same field-amplitude envelope. Every composition lacking the two Euler operators sits near $0.68$.

\paragraph{Scaling discovery to native $256^2$.}
Label-free adaptation from the $256^2$ library composes at $0.0079 \pm 0.0003$ over three seeds against the supervised $0.0031$, with both mechanisms identified on every seed. The full discovery loop improves on adaptation. At the registered budget (WIN$=8$, BS$=6$, $H\in\{1,2\}$, gradient-checkpointed RK4 with verified gradient equivalence), ten seeds give median composed $0.00164$ at the registered-primary selected round with $10/10$ identified; a secondary rollout selector reaches $0.00093$ at $7/10$ identified. The means overlap ($0.00279 \pm 0.00254$ label-free against $0.00317 \pm 0.00034$ supervised) and a per-seed sign test at $n{=}10$ ($7/10$) is not significant, so we carry the claim at the median with the spread stated. The two seeds that lose Euler at the final round share one collapse mode: the Euler operators drift onto the diffusion action. Run on coupled flows at native $256^2$, where no labeled data exists at any resolution, the loop identifies both mechanisms on ten of ten seeds and composes at $0.0116 \pm 0.0014$ (median $0.0118$).

\paragraph{Selecting the adaptation length without labels.}
Across rounds $\{1,2,4,8\}$, the held-out operator-fit residual on a disjoint unlabeled shard separates the identifying rounds (about $3.10$) from those where over-adaptation collapses the Euler operator into diffusion (about $4.76$), and selects round $1$ without seeing identification. The collapsed round $4$ composes best of all ($0.0038$) with only one identified mechanism, yet the rule still prefers the fully identified library.

\subsection{Robustness and scope}\label{app:robustness}

\paragraph{Noise.}\label{app:noise}
Under additive observation noise, PDE-FIND's utility degrades by a factor of $2.4$ by $10\%$ noise while the learned library degrades by $1.2$, so the gap narrows; PDE-FIND still leads on utility at every level (composed NRMSE $0.056 \to 0.134$ against $0.124 \to 0.150$ from no noise to $10\%$), and its selection of active terms stays more accurate ($0.91$ to $0.98$ against $0.41$ to $0.49$).

\paragraph{A shared encoder across dimensions.}\label{app:tierc}
The main-text libraries are per-dimension (\S\ref{sec:data}). We also trained a single shared-trunk encoder on five 1D mechanisms and three 2D mechanisms jointly with every criterion pre-registered. On all three seeds the shared encoder identifies all eight mechanisms, and rollout utility stays within the pre-registered $2\times$ of per-dimension baselines (1D $1.28/1.25/1.08\times$; 2D $0.98/1.99/1.01\times$, the $1.99$ marginal). Cross-dimension alignment fails, as pre-registered at $30\%$: codes of the same mechanism are $0/2$ mutual nearest neighbors across dimensions (permutation $p=1.0$) on all three seeds.

\subsection{External transfer to The Well}\label{app:well-figanchor}\label{sec:res-transfer}\label{app:well}

\paragraph{Zero-shot transfer.} We apply the synthetic 2D library, without fine-tuning, to the Well's \texttt{shear\_flow} vorticity channel (Reynolds $10^4$, Schmidt $1$) and \texttt{rayleigh\_benard} buoyancy channel (Rayleigh $10^6$, Prandtl $1$), at native resolution, with the same context length, beam harness, and search budget as Table~\ref{tab:nsfinal}, scored by the Well's VRMSE (Table~\ref{tab:well}). The published baselines are the Well's own; its exact rollout length is not pinned in the released configuration, so we report time-averaged VRMSE at 10 and 50 steps. The diffusion-selected trajectories average $0.52$ / $0.77$ (10 / 50 steps) and smooth toward the mean, keeping far more high-wavenumber energy than the true flow ($53\times$ at mid-horizon); the all-trajectory 50-step mean ($2.46$) is dominated by the single Euler-selected divergence.

\begin{table}[!htbp]
\centering\footnotesize
\caption{Transfer to The Well with no fine-tuning. VRMSE (predicting the mean scores $1.0$), time-averaged, 50-step horizon unless noted; baselines are trained on the dataset, our library never sees it.}
\label{tab:well}
\resizebox{\linewidth}{!}{%
\begin{tabular}{llll}
\toprule
Dataset & Beam selection & Ours (zero-shot) & Trained baselines \\
\midrule
\texttt{shear\_flow} (3/4 traj.) & diffusion & 0.77 & CNextU-Net 0.81; FNO 1.19; TFNO 1.47; U-Net 3.45 \\
\texttt{shear\_flow} (1/4 traj.) & Euler & 7.5 (1.39 at 10 steps) & \\
\texttt{rayleigh\_benard} & none engaged & declines to compose & \\
\bottomrule
\end{tabular}}
\end{table}

Zero-shot discovery on \texttt{shear\_flow} identifies diffusion ($0.0009$ to $0.0015$ against a null of $0.014$) but not Euler.

\paragraph{Label-free adaptation on shear flow (pre-registered).} Adaptation on 32 unlabeled shear-flow realizations disjoint from the graded trajectories recovers the advective mechanism: adapted discovered operators identify as Euler on three of three seeds, where the zero-shot discovered operators had missed the threshold by margins as small as $0.0005$. Composed transfer improves on a matched base ($0.868$ to $0.818$, more than twice the seed spread) but misses the pre-registered $0.72$ bar, and the Euler-selected trajectory still diverges: adaptation identifies the mechanism but does not improve composed transfer to the registered level. The bar was set against a different base model than the adaptation harness used, so only the same-base change is quoted.

\paragraph{Developed turbulence.}\label{app:trl2d} On developed turbulence from The Well \citep{ohana2024well}, the beam selects single-operator rollouts and never mixes operators within a step, so composition provides no benefit over a monolithic rollout. We read this as a boundary of the splitting paradigm rather than of discovery: at developed-turbulence scales the mechanisms are not weakly coupled, and the local $O(\Delta t^3)$ splitting error of \S\ref{sec:background} ceases to be small.


\section{Growing a mechanism absent from pretraining}
\label{app:reach}

\paragraph{Setup.} The backbone is pretrained on all mechanisms except Euler. Growth data mix Euler trajectories with diffusion trajectories, which the backbone already knows. All errors are relative $L^2$ on held out windows from one disjoint split; ranges are over three seeds.

\paragraph{New operator.} The new operator has the architecture of a library operator at width 16, starts from a small random initialization, and is trained with its weight fixed to one; the assignment weights of Eq.~\eqref{eq:wstep} are solved only at evaluation. With a zero initialization, the weight step sets the new weight to zero before the operator trains, and it never recovers. The pretrained weights are checked bitwise unchanged at every checkpoint.

\paragraph{Label free routing.} A window is routed to the new operator when the frozen backbone's one step error on it exceeds the $95$th percentile of the same error over the diffusion windows (threshold $0.00101$). The decision is per window. Recall on Euler windows is $1.0$ and precision $0.94$ to $0.95$; the false positives are the top $5\%$ of diffusion windows, as the threshold implies.

\paragraph{Identification.} A standalone operator does not share the amplitude normalization of operators inside the library, so its fingerprint is computed with a fitted amplitude, and the nulls (zero operator and eight random operators) are rescored the same way. The fully retrained Euler code serves as positive control ($0.00063$); the new operators score $0.00055$ to $0.00063$ against a null of $0.00290$.

\paragraph{Variants.} Routing by labels instead of residuals gives a slightly lower one step error ($0.053$ to $0.058$) but a higher step ten error ($0.39$ to $0.43$). Training the new operator on three step rollouts lowers its step ten error to $0.28$ to $0.31$ at three GPU hours. Retraining without replay forgets diffusion (its error grows about sixty times) and leaves identification invalid. Full retraining reaches $0.038$ at step one and $0.127$ at step ten, at about twenty times the compute.

\paragraph{Rollouts.} Table~\ref{tab:reach-full} reports steps 1 and 10; at steps 3 and 5 the ordering is the same (new operators $0.12$ to $0.19$, replay $0.11$ to $0.14$).

\begin{table}[t]
\centering\small
\caption{Adding Euler to a network pretrained without it. \emph{Euler}: identified seeds (\S\ref{sec:identify}). Errors are composed relative $L^2$ on held out Euler windows at rollout steps 1 and 10; diffusion error is relative to the original library. Ranges over three seeds. New operators have width 16 and are trained on single steps.}
\label{tab:reach-full}
\setlength{\tabcolsep}{4pt}
\begin{tabular}{lccccc}
\toprule
Arm & Euler & Step 1 & Step 10 & Diffusion & GPU h \\
\midrule
Frozen, no growth & no & $0.221$ & $0.88$ & $1.00\times$ & 0 \\
Retrain, with replay & 2/3 & $[0.061, 0.073]$ & $[0.19, 0.22]$ & $[1.06, 1.42]\times$ & 9 \\
Retrain, no replay & invalid & n/a & n/a & $[60, 67]\times$ & 9.5 \\
New operator, labels & 3/3 & $[0.053, 0.058]$ & $[0.39, 0.43]$ & $1.00\times$ & 1.5 \\
New operator, label free & 3/3 & $[0.057, 0.060]$ & $[0.32, 0.34]$ & $1.00\times$ & 1.5 \\
\midrule
Full retraining & yes & $0.038$ & $0.127$ & n/a & 29 \\
\bottomrule
\end{tabular}
\end{table} 
\section{Reproducibility and NOS Comparison}
\label{app:repro}\label{app:nos}

\subsection{NOS reference and our reproduction}
\label{app:repro:nos}\label{app:repro:ours}

\paragraph{NOS reference.} We evaluated the NOS authors' released checkpoint with their released evaluation code, unmodified except for a checkpoint-loading shim. On the released \nsD{} test data this gives a composed zero-shot NRMSE of $0.0045$ and a direct rollout of $0.021$ at $128^2$ (Table~\ref{tab:nsfinal}). The headline NRMSE printed in the NOS paper ($0.066$) is not reproduced by the released code on the released data; candidate causes are an evaluation-split difference and the evaluation resolution ($256^2$ there, $128^2$ here). Our comparisons are made against what the released artifacts produce under one shared protocol, never against the printed number. Reproducing the NOS numbers required the \texttt{nos} repository at commit \texttt{6adbdcd}, its vendored \texttt{torchdiffeq}, and \texttt{typeguard}~$<$3.

\paragraph{Our reproduction.} Our supervised operator uses the NOS architecture, its RelativeL2 loss, its NS configuration, and its rollout-horizon curriculum (horizon ramped from 1 to 16). It omits the VQ-codebook loss and in-context learning, and fixes the integration substeps at 4. To check that these omissions do not matter, we also retrained the full method (codebook, in-context, substep ramp; 23M parameters) on the full corpus of $14{,}746$ mechanism trajectories used for the released checkpoint. It composes NS at $0.0053$, in the range of the released checkpoint ($0.0045$) and our NOS operator ($0.0036\pm0.0006$). Within one codebase, corpus, and schedule ($20{,}000$ steps), switching the codebook and in-context learning off moves the composed NRMSE from $0.0053$ to $0.0060$. At native $256^2$ ($n{=}32$, shared harness), the released checkpoint composes at $0.0056$ and the retrained full method at $0.0043$; NOS's published $256^2$ direct rollout ($0.0016$) is a different evaluation mode. Training the 2D operator ($23.61$M parameters) takes about two GPU-hours on one consumer GPU (estimated).

\paragraph{Re-measured 1D battery.}\label{app:repro:audit} A September re-run of the 1D battery did not reproduce two earlier single-run NOS-recipe values ($0.036$ NOS operator, $0.060$ full NOS method), which re-measure at $0.096 \pm 0.005$ and $0.107 \pm 0.009$ over three seeds, while the control row ($0.096 \pm 0.001$) and the monolithic baseline ($0.117 \pm 0.001$) reproduced exactly. We could not identify the cause. The printed tables use only the re-measured values, and every result file carries a provenance record (checkpoint path and modification time, dictionary, code date, library versions).

\subsection{Resolution transfer without retraining}
\label{app:restransfer}

\begin{table}[!htbp]
\centering\small
\caption{Zero-shot Navier--Stokes on the released \nsD{} test trajectories ($n{=}32$; per-step-mean NRMSE over the released 10-step horizon; one protocol for all models). The slice evolves slowly, so every row is read against the persistence floor. ``Composed'': operators trained on pure mechanisms only, combined by beam-searched Strang splitting. The $256^2$ column holds models trained or adapted at native resolution. Rows are mean$\pm$sd where multi-seed; NOS is its single released checkpoint, the retrained full method a single run; the claim is parity. The $256^2$-trained NOS-operator row (3 seeds) lands below both references; with single-run references the claim is parity, not superiority. $^{\dagger}$the full label-free loop of \S\ref{sec:method-ssl} at native $256^2$, run at its registered budget: median composed over 10 seeds at the pre-registered label-free-selected round, both mechanisms identified on 10 of 10 seeds at that round; 7 of 10 seeds individually beat their supervised warm start. A secondary rollout-based selector reaches median $0.00093$ but identifies both mechanisms on only 7 of 10. A lighter adaptation-only variant reaches $0.0079 \pm 0.0003$ (Appendix~\ref{app:disc2d}). }
\label{tab:nsfinal}
\resizebox{\linewidth}{!}{%
\begin{tabular}{lcc}
\toprule
Model & Composed, $128^2$ & Composed, $256^2$ \\
\midrule
persistence (repeat last frame) & 0.012 & 0.017 \\
\midrule
\multicolumn{3}{l}{\emph{supervised (labeled pure-mechanism trajectories)}} \\
released NOS checkpoint \citep{serrano2026test} & 0.0045 & 0.0056 \\
NOS operator, this work (codebook, in-context omitted) & 0.0036 $\pm$ 0.0006 & --- \\
NOS operator, trained at $256^2$ (this work) & --- & 0.0031 $\pm$ 0.0004 \\
full NOS method, retrained (this work) & 0.0053 & 0.0043 \\
\midrule
\multicolumn{3}{l}{\emph{discovered, label-free (ours: the method of \S\ref{sec:method-ssl}, from unlabeled mixtures)}} \\
\dlib{} from pure-trajectory mixtures (\S\ref{sec:res-nsssl}) & 0.0038 $\pm$ 0.0001 & --- \\
\dlib{} from coupled NS (\S\ref{sec:res-nsssl}) & $0.056$ & $0.0116 \pm 0.0014$ (10 seeds) \\
\dlib{} full discovery at $256^2$, registered budget, 10 seeds & --- & $0.0028 \pm 0.0025$, median $0.0016$ (10/10 identified)$^{\dagger}$ \\
\bottomrule
\end{tabular}}
\end{table}
Every $128^2$-trained library also runs at $256^2$ with no retraining; the cost is roughly uniform across constructions: the supervised NOS operator moves from $0.0036$ to $0.0090$, the discovered library from pure-trajectory mixtures from $0.0038$ to $0.013$, from $0.0038$ to $0.013$ (released 10-step protocol, same harness as Table~\ref{tab:nsfinal}). Training or adapting at native resolution removes this cost (the $256^2$ rows of Table~\ref{tab:nsfinal}). 
Direct, composition-free rollouts at $128^2$ on the same protocol: released NOS checkpoint $0.021$ ($0.0016$ at native $256^2$, its published mode), our NOS operator $0.0038 \pm 0.0006$, the retrained full NOS method $0.052$, the only entry above the persistence floor.

\end{document}